\documentclass{article}
\usepackage[margin=1in]{geometry}

\usepackage[round]{natbib}

\usepackage{std_definitions}

\usepackage{amsfonts}
\usepackage{nicefrac}
\usepackage{xspace}
\usepackage{enumitem}

\usepackage{caption}

\usepackage{algorithm,algorithmicx}
\usepackage{setspace}
\usepackage[noend]{algpseudocode}
\usepackage{optidef}
\usepackage{soul}

\usepackage{cleveref} 
\newtheorem{assumption}{Assumption}
\Crefname{assumption}{Assumption}{Assumptions}

\newcommand{\algname}{\mbox{SABRE}\xspace}

\newcommand{\prob}[0]{\textup{Prob\xspace}}

\newcommand{\pisafe}{\Pi_{\textup{safe}}\xspace}
\newcommand{\piknown}{\pi_{\textup{safe}}\xspace}
\newcommand{\asafe}{a_{\textup{safe}}\xspace}
\newcommand{\rd}{\textup{RD}\xspace}
\newcommand{\thetamin}{\theta_{\textup{min}}\xspace}
\newcommand{\dnnr}{d_{\textup{NNR}}\xspace}

\newcommand{\poly}{{\tt Poly}}

\newcommand{\SubOpt}{{\tt SubOpt}}

\newcommand{\sign}{\textup{sign}\xspace}

\newcommand{\epsexplore}{\epsilon_{\textup{explore}}\xspace}
\newcommand{\epsfinal}{\epsilon_{R}\xspace}
\newcommand{\deltaexplore}{\delta_{\textup{explore}}\xspace}
\newcommand{\deltafinal}{\delta_{R}\xspace}
\newcommand{\vcf}{d_{\textup{VC}}\xspace}
\newcommand{\pcd}{d_{\Pi}\xspace}
\newcommand{\thetamax}{d_{\theta}\xspace}

\newcommand{\Alg}{{\tt Alg}}

\newcommand{\mycomment}[1]{\hfill\textcolor{blue}{//~#1}}

\begin{document}

\title{Provable Safe Reinforcement Learning with Binary Feedback}

\author{Andrew Bennett \\ Cornell University \\ awb222@cornell.edu \and Dipendra Misra \\ Microsoft Research \\ dimisra@microsoft.com \and Nathan Kallus \\ Cornell University \\ kallus@cornell.edu}

\date{October 2022}

\maketitle

\begin{abstract}
Safety is a crucial necessity in many applications of reinforcement learning (RL), whether robotic, automotive, or medical.
Many existing approaches to safe RL rely on receiving numeric safety feedback, but in many cases this feedback can only take binary values; that is, whether an action in a given state is safe or unsafe. 
This is particularly true when feedback comes from human experts. We therefore consider the problem of provable safe RL when given access to an offline
oracle providing binary feedback on the safety of state, action pairs. We provide a novel meta algorithm, $\algname$, which can be applied to any MDP setting given access to a blackbox PAC RL algorithm for that setting. $\algname$ applies concepts from active learning to reinforcement learning to provably control the number of queries to the safety oracle. $\algname$ works by iteratively exploring the state space to find regions where the agent is currently uncertain about safety. Our main theoretical results shows that, under appropriate technical assumptions, $\algname$ never takes unsafe actions during training, and is guaranteed to return a near-optimal safe policy with high probability. We provide a discussion of how our meta-algorithm may be applied to various settings studied in both theoretical and empirical frameworks.
\end{abstract}

\section{Introduction}

Reinforcement learning (RL) is an important paradigm that can be used to solve important dynamic decision-making problems in a diverse set of fields, such as robotics, transportation, healthcare, and user assistance. In recent years there has been a significant increase in interest in this problem, with many proposed solutions. However, in many such applications there are important safety considerations that are difficult to address with existing techniques.

Consider the running example of a cleaning robot, whose task is to learn how to vacuum the floor of a house. The primary goal of the robot, of course, is to learn to vacuum as efficiently as possible, which may be measured by the amount cleaned in a given time. However, we would also like to impose safety constraints on the robot's actions; for example, the robot shouldn't roll off of a staircase where it could damage itself, it shouldn't roll over electrical cords, or it shouldn't vacuum up the owner's possessions. In this example, there are several desirable properties we would like a safety-aware learning algorithm to have, including:
\begin{enumerate}
    \item The agent should avoid taking any unsafe actions, \emph{even during training}
    \item Since it is hard to concretely define a safety function from the robot's sensory observations \emph{a priori}, we would like the agent to \emph{learn} a safety function given feedback of observed states
    \item Since the notion of safety is human-defined,
    and we would like the safety feedback to be manually provided by humans (\emph{e.g.} the owner), we would want the agent to ask for \emph{as little feedback as possible}
    \item We would like to use \emph{binary} feedback (\emph{i.e.} is an action in a given state safe or unsafe) rather than numeric feedback, as this is more natural for humans to provide
    \item Since the agent may need to act in real time without direct intervention, they should only ask for feedback \emph{offline} in between episodes 
\end{enumerate}

Moving away from our specific example, the above five properties would be ideal for safe RL in many applications where safety is naturally human-defined. Unfortunately, there are no existing safe RL methods that can provably satisfy all of these properties. In particular, existing safe RL methods all fail to satisfy these properties for at least one of the following reasons: (1) they assume a safety function is fully known (\emph{e.g.} \citealp{chow2018lyapunov,simao2021alwayssafe}); (2) they learn safety using numeric feedback (\emph{e.g.} \citealp{wachi2020safe,amani2021safe}); or (3) they require a human-in-the-loop who can intervene and monitors the safety of every action taken in real time \citep{saunders2018trial}. In addition, none of these existing methods address the issue of only asking for minimal feedback, which is an important consideration which places the problem at the interface of RL and active learning.

In this paper, we present a novel safe RL framework encompassing the above concerns, involving an offline safety-labeling oracle that can provide binary feedback on the safety of state, action pairs in between episodes. We provide a meta algorithm, \algname, which can be applied to \emph{any} MDP class within this safety framework, given a blackbox RL algorithm for the MDP class. This algorithm utilizes ideas from disagreement-based active learning, by iteratively exploring the state space to find all regions where there is disagreement on possible safety. Under some appropriate technical assumptions, including that the blackox RL algorithm can provably optimize any given reward, \algname will satisfy all of the above five properties, and will return an approximately optimal safe policy with high probability. Importantly, we provide high-probability bounds on the number of samples and calls to the labeling oracle needed, and show that they are both polynomial, and that the latter is lower order than the former. Finally, we provide some discussion of how this meta-algorithm approach may be applied in various settings of both theoretical and practical interest.

\paragraph{Math Notation} For any natural number $n \in \NN$, we let $[n]$ denote the set $\{1, 2, \cdots, n\}$. For any countable set $\Xcal$, we let $\Delta(\Xcal)$ denote the space of all probability distributions over $\Xcal$. Given sets $\Xcal$ and $\Ycal$, we let $\Xcal \to \Ycal$ denote the set of all functions from $\Xcal$ to $\Ycal$. Lastly, $\sign(y)$ denotes the sign function which takes a value of 1 if $y > 0$, a value of 0 at $y=0$, and $-1$ if $y < 0$.

\section{Problem Setup}

\paragraph{Markov Decision Process} We consider learning in finite-horizon Markov decision processes (MDPs). A \emph{reward-free} MDP is characterized by a tuple $(\Scal, \Acal, T, \mu, H)$, where $\Scal$ is a given (potentially infinitely large) state space, $\Acal$ is a finite action space with $|\Acal| = A$, $T: \Scal \times \Acal \to \Delta(\Scal)$ is a transition operator, $\mu \in \Delta(\Scal)$ is an initial state distribution, and $H \in \NN$ is the horizon.
An MDP (with reward) is a tuple $(\Scal, \Acal, T,R, \mu, H)$, where $R: \Scal \times \Acal \rightarrow [0, 1]$ is the reward function.
We assume that the transition operator and reward function are time homogeneous; \emph{i.e.}, $T$ and $R$ do not depend explicitly on the time index.\footnote{This is w.l.o.g. since we can always include the current time index in the observation definition.}
We will let $(M,R)$ refer to the MDP we are learning in, where $M$ is the corresponding reward-free MDP.

The agent interacts with an MDP over a series of rounds. In each round the agent successively takes actions according to some \emph{policy}, which is a mapping from states to actions, in order to generate an \emph{episode}. Specifically, in the $n$'th round for each $n \in \NN$, the agent selects some policy $\pi_n \in \Scal \to \Delta(\Acal)$, which it then uses in order to generate an episode $\tau_n = (s^n_1,a^n_1,r^n_1,\ldots,s^n_H,a^n_H,r^n_H,s^n_{H+1})$, where $s^n_1 \sim \mu(\cdot)$, and for each $h \in [H]$ we have $a^n_h \sim \pi_n(\cdot \mid s^n_h)$, $r^n_h = R(s^n_h,a^n_h)$, and $s^n_{h+1} \sim T(\cdot \mid s^n_h,a^n_h)$.

For any given policy $\pi$, we let $\EE_\pi[\cdot]$ and $\PP_\pi(\cdot)$ denote the expectation and probability operators over trajectories generated using $\pi$ in the MDP $(M,R)$. Also, for any policy $\pi$, we define the \emph{value function} $V^\pi \in \Scal \to \RR$ according to $V^\pi(s) = \EE_\pi[\sum_{h=1}^H R(s_h, a_h) \mid s_1=s]$, and we similarly define the value of the policy according to $V(\pi) = \EE_{s \sim \mu(\cdot)}[V^\pi(s)]$. 

Our main metric of success for reinforcement learning is \emph{suboptimality} ($\SubOpt$). For any policy class $\Pi \subseteq \Scal \to \Delta(\Acal)$ and any policy $\hat\pi \in \Pi$, we define $\SubOpt(\hat\pi; \Pi) = \sup_{\pi \in \Pi} V(\pi) - V(\hat\pi)$. Then, one of our primary goals is to learn a policy $\hat \pi$ that has low suboptimality with respect to a given set of policies that it must choose from, in a relatively small number of episodes.

Over the past few decades, many reinforcement learning algorithms have been developed that can provably obtain low suboptimality using a small number of episodes, for various classes of MDPs~\cite{}. Our focus is on adapting such RL algorithms, in order to additionally take into account safety considerations. For these reasons, we take a meta-learning approach, and assume access to a blackbox reinforcement learning algorithm $\Alg$. We formalize this as follows: 

\begin{assumption}
\label{assum:blackbox-rl}
    We have access to a blackbox RL algorithm $\Alg$ for a set of reward-free MDPs $\Mcal$ that contains $M$. Specifically, given \emph{any} reward function $R' : \Scal \times \Acal \to [0,1]$ and policy class $\Pi$ as input, along with any $\epsilon,\delta \in (0,1)$, $\Alg(R',\Pi,\epsilon,\delta)$ returns a policy $\hat\pi \in \Pi$ such that $\SubOpt(\hat\pi;\Pi) \leq \epsilon$ with probability at least $1-\delta$. Furthermore, it is guaranteed to do so after at most $n_{\Alg}(\epsilon,\delta)$ episodes, for some $n_{\Alg}(\epsilon, \delta) = \poly(\epsilon, \log(1/\delta))$, while only following policies in $\Pi$.
\end{assumption}

We refer to $n_{\Alg}(\epsilon,\delta)$ as the \emph{sample complexity} of $\Alg$, and note that it implicitly depends on the time horizon $H$. We also note that the requirement that $\Alg$ only follows policies in $\Pi$ is important, as in practice we will call $\Alg$ using classes of policies that are guaranteed to be safe.

\begin{figure*}[t]
    \centering
    \hfill\includegraphics[width=0.45\textwidth]{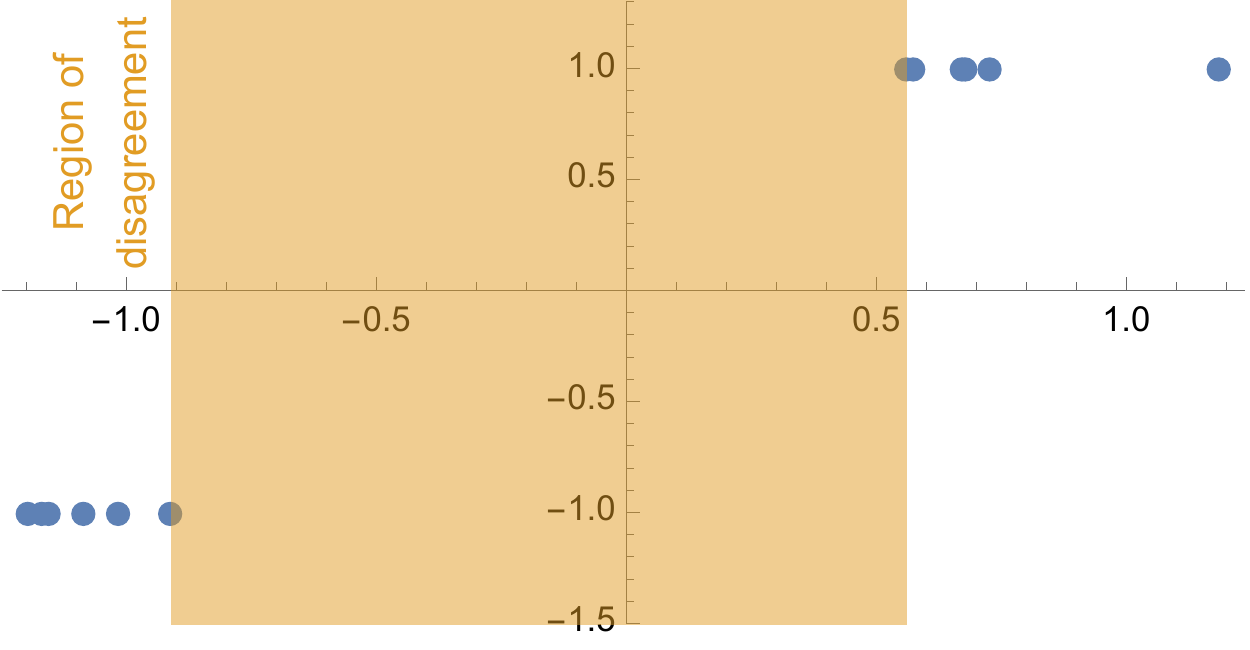}\hfill
    \includegraphics[width=0.45\textwidth]{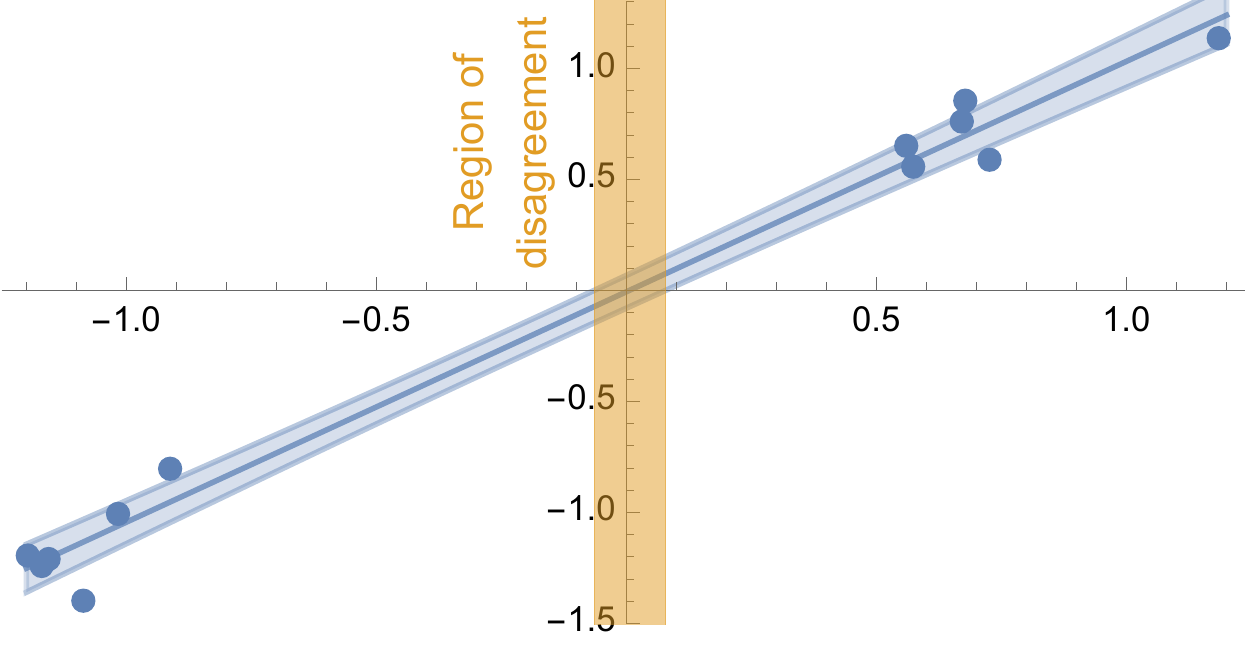}\hfill
    \caption{Illustrating the challenge with binary safety feedback. \textbf{Left}: we observe binary safe/unsafe labels (our setting) and we know the safety function is a halfspace; the safety of state-action pairs in the region of disagreement (shaded gold region) is uncertain, as there exist halfspaces consistent with both the observed data and these being either safe or unsafe. \textbf{Right}: we observe noisy numeric safety values (positive being safe) and we know their mean is linear; the region of disagreement (using 95\% confidence) is far smaller because we can extrapolate to yet-unseen state-action pairs.
    }
    \label{fig:binary-safety}
\end{figure*}

\paragraph{Safety Considerations}
In addition to the general MDP setup presented above, we would also like to take
safety considerations into account. Specifically, we will use a binary notion of safety as follows: we assume there exists a function $f^\star: \Scal \to \{\pm 1\}$ such that $f^\star(s,a) = 1$ means that taking action $a$ in state $s$ is safe, and otherwise action $a$ in state $s$ is unsafe. Importantly, we assume that the safety function $f^\star$ is \emph{unknown},
and we require the agent to only take actions $a$ in states $s$ such that $f^\star(s,a) = 1$, \emph{even during training}. That is, the agent must \emph{learn} the safety relation, while simultaneously maintaining safety.

The assumption that the safety values are binary is in contrast to most of the literature, which mostly assumes that these values are numeric (\emph{i.e.} takes values in $\RR$ rather than $\{\pm 1\}$, with safety given by $\sign(f^\star(s,a))$; see \pref{sec:related-work} for a detailed discussion).
To see why this makes the problem more challenging, consider for example the example displayed in \pref{fig:binary-safety}. In this example, we consider the case where $\Scal = \RR$, and the safety function is assumed to be linear. On the left we display a series of observed binary safety values for some fixed action, and on the right we display the corresponding observations when the safety values are numeric. In each case, we shade in gold the \emph{disagreement region} where we are unsure about safety values. With the numeric observations, we can be sure of safety for almost all states except near the $f^\star(s,a) = 0$ margin, whereas with the binary observations we are unsure of safety for \emph{all} states in between the two observed clusters. Note that this is true even though the numeric observations have noise, while the binary ones don't.
In other words, when we receive binary safety feedback, we cannot easily extrapolate safety beyond the observed states.

In order to make the task of learning given these safety constraints feasible, we model $f^\star$ using some class $\Fcal \subseteq \Scal \times \Acal \to \{\pm 1\}$ of candidate safety functions. This allows us to ensure that $f^\star$ is learnable, by using a sufficiently well-behaved class $\Fcal$. In particular, in our theory we will focus on classes $\Fcal$ with finite Vapnik-Chervonenkis (VC) dimension \citep{dudley1987universal}. 
Then, in order to make it feasible to guarantee safety during training, we make the following core assumption on the setting.

\begin{assumption}
\label{assum:safety-realizability}
    The safety class $\Fcal$ is correctly specified; that is, $f^\star \in \Fcal$. Furthermore, there exists a policy $\piknown \in \Pi$ that is known to always take safe actions.
\end{assumption}
To explain the importance of this assumption, let us define the set of safe policies corresponding to each $f \in \Fcal$, by
\begin{equation*}
    \pisafe(f) = \{\piknown\} \cup \{\pi \in \Pi : f(s, \pi(s)) = 1~\textup{a.s.}~\forall s \in \Scal \} \,.
\end{equation*}
That is $\pisafe(f)$ is the set of all policies that would be safe if $f$ was the true safety function, along with the known safe policy. We similarly define the shorthand $\pisafe = \pisafe(f^\star)$ for the actual set of safe policies.
Then, the assumption that $\Fcal$ is correctly specified allows us to reason about which policies are safe, since if $\pi \in \pisafe(f)$ for all possible $f \in \Fcal$ that are consistent with our observations so far, we are guaranteed that $\pi \in \pisafe$.
Furthermore, the existence of a known safe policy $\pisafe$ allows the agent to avoid getting stuck with no known safe action to perform.
Note that the definition of $\piknown$ could be problem dependent. For example, in the previous cleaning robot example, the policy that always takes the ``don't move'' action could qualify. Alternatively, in different applications, $\piknown$ could be defined for example by taking a special action that terminates the episode early, or has an expert take control.

\paragraph{Obtaining Safety Feedback}
A key motivation behind using binary safety feedback rather than continuous is for applications where safety is human-defined. In such applications, obtaining safety labels for $(s,a)$ pairs may be expensive or otherwise burdensome. 
This motivates an \emph{active learning}-style approach to the problem, where we have to explicitly ask for labels of $(s,a)$ pairs, with an additional goal of minimizing the number of times we do so. 

Concretely, our model for the labeling process is as follows. We assume access to a labeling oracle, which can be given $(s,a)$ pairs and returns the corresponding values of $f^\star(s,a)$. We assume that this oracle can be queried an unlimited number of times in between episodes, and that it can only be queried with states $s$ that have been previously observed. The reason for the latter restriction is that in many applications the state corresponds to some kind of observation of the environment, such as an image, and therefore the total space of states $\Scal$ is not a-priori known.

\paragraph{Problem Setup Summary.} Given a class of MDPs $\Mcal$, a policy class $\Pi$, a blackbox RL algorithm $\Alg$, a safety function class $\Fcal$, a desired sub-optimality $\epsilon$, and failure probability $\delta$, we want to propose an online learning algorithm that, for any unknown $M \in \Mcal$ and $f^\star \in \Fcal$, interacts with the MDP over $N$ rounds and returns a policy $\hat\pi \in \Pi$ such that, with probability at least $1-\delta$, we have:
\begin{enumerate}
    \item $\SubOpt(\hat\pi;\Pi) \leq \epsilon$ ($\hat\pi$ is approximately optimal)
    \item $\hat\pi \in \pisafe$ (the returned policy is safe)
    \item $f^\star(s_h^n,a_h^n) = 1$ for all $h \in [H]$ and $n \in [N]$ (the agent never takes unsafe actions during training)
\end{enumerate}
In addition, we would like to establish sample-complexity bounds on the number of episodes $N$ needed to ensure the above, in terms of $1/\epsilon$, $\log(1/\delta)$, $H$, and $n_{\Alg}(\cdot)$. Finally, we would like to establish corresponding high-probability bounds on the total number of calls to the labelling oracle, which are lower order than the total sample complexity.

\section{Learning Safety via Active Learning in Reinforcement Learning}
\label{sec:learn-safety}

First, we establish some basic definitions and a result that will form the basis for safety learning.
For any safety-labeled dataset $\Dcal$ consisting of $(s, a, f^\star(s,a))$ tuples, we define the corresponding \emph{version space} of safety functions consistent with $\Dcal$ by
\begin{equation*}
    \Vcal(\Dcal) = \{f \in \Fcal : f(s,a) = c \quad \forall (s, a, c) \in \Dcal \} \,.
\end{equation*}
Similarly, for any given $a \in \Acal$, we can define the corresponding \emph{region of disagreement} of states where safety of that action is not known given $\Dcal$ by
\begin{equation*}
    \rd^a(\Dcal) = \{s : \exists f,f' \in \Vcal(\Dcal)~\textup{s.t.}~f(s,a) \neq f'(s,a)\} \,.
\end{equation*}
Note that these are both standard definitions in disagreement-based active learning \citep{hanneke2014theory}.

Next, define the set of policies known to surely be safe given $\Dcal$ as follows:
\begin{equation*}
    \Pi(\Dcal) = \bigcap_{f \in \Vcal(\Dcal)} \pisafe(f) \,.
\end{equation*}
Note that given \pref{assum:safety-realizability}, $\Pi(\Dcal)$ is always ensured to be non-empty, as it will at least contain $\piknown$. 

Given these definitions, we have the following lemma.

\begin{lemma}
\label{lem:subopt-bound}
    For any safety labeled dataset $\Dcal$, let
    \begin{equation*}
        U(\Dcal) = \sup_{\pi \in \pisafe} \PP_\pi(\exists h \in [H], a \in \Acal : s_h \in \rd^a(\Dcal)) \,.
    \end{equation*}
    Then, for any $\Dcal$ and $\hat\pi \in \Pi(\Dcal)$, we have 
    \begin{equation*}
        \SubOpt(\hat\pi;\pisafe) \leq \SubOpt(\hat\pi;\Pi(\Dcal)) + H U(\Dcal) \,.
    \end{equation*}
\end{lemma}

\subsection{Challenges with a Naive Approach}
\label{sec:baseline-approach}

This lemma in fact suggests a rather simple, naive approach, based on greedily trying to follow the blackbox RL algorithm, and switching to $\piknown$ and querying the safety oracle whenever we reach a state where safety is not known.
Before we outline our novel solution that addresses the challenges outlined in the previous section, we consider the limitations of this naive approach.

Given \pref{lem:subopt-bound} and \pref{assum:blackbox-rl}, we can easily ensure sub-optimality of at most $\epsilon + H U(\Dcal)$ for any target $\epsilon$ using a naive approach like above, where $\Dcal$ is the labeled dataset incidentally collected following this approach. However, since this approach does nothing explicit to try to learn the safety and shrink the region of disagreement, it is difficult to provide any guarantees on how fast $U(\Dcal)$ shrinks.

A second issue is that this approach provides no control on the number of calls to the labelling oracle.
Existing analyses from disagreement-based active learning provide bounds on the number of samples needed to shrink regions of disagreement under fixed distributions, but the agent may roll out with a different distribution every episode.

\subsection{The \algname Algorithm}

\begin{algorithm*}[t]
\setstretch{1.1}
\renewcommand{\algorithmicrequire}{\textbf{Input:}}
\renewcommand{\algorithmicensure}{\textbf{Output:}}
\caption{SAfe Binary-feedback REinforcement learning (\algname)}
\label{alg:ours}
\begin{algorithmic}[1]
\Require{Number of epochs $N$, number of iterations per epoch $B$, number of rollouts per batch $m$, accuracy parameters $\epsexplore$ and $\epsfinal$, and probability parameters $\deltaexplore$ and $\deltafinal$, initial safety-labeled dataset $\Dcal_0^{(B)}$ (optional)}
\Ensure{Final policy $\hat\pi$}
    \For{$n = 1,2,\ldots,N$}
        \State $\Pi_n \leftarrow \Pi(\Dcal_{n-1}^{(B)})$ \mycomment{define a safe policy class for the current safety labeled dataset $\Dcal_{n-1}^{(B)}$}
        \State $\Dcal^{(0)}_n \gets \Dcal_{n-1}^{(B)}$
        \For{$i = 1,2,\ldots,B$}
            \State $\widetilde R_n^{(i)}(\cdot) \leftarrow \one\{\exists a \in \Acal : \cdot \in \rd^a(\Dcal^{(i-1)}_n)\}$ \mycomment{safety reward that incentivizes visiting RD}
            \State $\hat \pi_n^{(i)} \gets \Alg(\widetilde R_n^{(i)}, \Pi_n, \epsexplore, \deltaexplore)$\mycomment{call blackbox RL method with safety exploration reward} 
            \State Roll out with $\hat\pi_n^{(i)}$ for $m$ episodes, and collect all observed states in $\Scal_{n}^{(i)}$
            \State Expand $\Dcal_n^{(i-1)}$ to $\Dcal_n^{(i)}$ by labelling all $s \in \Scal_n^{(i)}$ and $a \in \Acal$ such that $s \in \rd^a(\Dcal_n^{(i-1)})$ \label{line:data_label}
        \EndFor
    \EndFor
    \State \Return $\Alg(R,\Pi(\Dcal_N^{(B)}),\epsfinal,\deltafinal)$ \mycomment{return a safe policy by optimizing the environment reward $R$}
\end{algorithmic}
\end{algorithm*}

Motivated by \pref{lem:subopt-bound}, as well as the limitations of the above naive baseline approach, we now present our novel algorithm in \pref{alg:ours}. This algorithm is superficially similar to the baseline approach, in the sense that it builds a labeled dataset $\Dcal$ over a series of rounds, and then estimates an optimal policy following $\Alg(R,\Pi(\Dcal),\epsilon,\delta)$. However, the difference is in how the safety-labeled dataset $\Dcal$ is constructed. Instead of successively trying to optimize the environmental reward, and labeling the states we happen observe in the process, our algorithm uses a strategy for constructing $\Dcal$ that explicitly targets $U(\Dcal)$.

Our algorithm uses two loops to construct the safety-labeled dataset. The outer loop runs over $N$ epochs. In each epoch it starts with the set $\Pi_n$ of policies known to be safe at the start of the epoch, and holds this set \emph{fixed} over the entire epoch. Within each epoch, the inner loop performs $B$ iterations, alternating between the following two steps: (1) (approximately) optimize a policy within $\Pi_n$ for hitting the region of disagreement with the current $\Dcal$; and (2) roll out with this policy for $m$ episodes to collect additional labeled data to expand $\Dcal$ with. The reason for having these two separate loops is that it allows us to derive our formal guarantees, as will be clear in the next section. However, it is possible that an improved analysis in future work could allow the algorithm to be simplified to a single loop.

Comparing with the naive approach discussed above, this algorithm follows a strategy for expanding $\Dcal$ based on actually optimizing hitting the region of disagreement for data collection, which is better tailored to explicitly reducing $U(\Dcal)$. Furthermore, the data that is used for expanding $\Dcal$ within each epoch is collected by rolling out with \emph{fixed} policies, which allows stronger control on the number of times the safety labeling oracle will be called.

\section{Theoretical Analysis}

We now provide a theoretical analysis of our proposed algorithm. First, by design \algname only ever takes safe actions during training, and the final returned policy $\hat\pi$ is always guaranteed to be safe. This holds with certainty, not just with high probability, since by definition $\Pi(\Dcal)$ contains only safe policies, for \emph{any} obtainable $\Dcal$. Given this, we will focus on establishing approximate suboptimality, along with high-probability bounds on the sample and labeling complexities.

Before we give our result, we must give some additional technical assumptions. First, we require that the MDP $M$ has reasonably low complexity, as follows.
\begin{assumption}
\label{assum:policy-cover}
    There exists some positive integer $\pcd$ such that, for any given set of policies $\Pi \subseteq \pisafe$, there exists a set of policies $\pi_1,\ldots,\pi_{\pcd}$ satisfying
    \begin{equation*}
        \frac{1}{H} \sum_{h=1}^H \PP_\pi(s_h \in \widetilde\Scal) \leq \sum_{i=1}^{\pcd} \left( \frac{1}{H} \sum_{h=1}^H \PP_{\pi_i}(s_h \in \widetilde\Scal) \right) \,,
    \end{equation*}
    for all measurable $\widetilde\Scal \subseteq \Scal$ and $\pi \in \Pi$.
\end{assumption}
We call the smallest $\pcd$ for which this assumption holds the \emph{policy-cover dimension} of the MDP. We give examples of $\pcd$ for some common MDP classes in \pref{sec:examples}.

In addition, we need need to assume a bound on the \emph{disagreement coefficient} of the distribution induced by any policy, which is a standard joint complexity measure of both distribution and hypothesis class, and is used to derive upper and lower bounds for active learning in binary classification \citep{hanneke2014theory}. Formally, for any $h \in [H]$ and $f,f' \in \Fcal$, let $\Ecal_h(f,f')$ denote the event that $f(s_h, a) \neq f'(s_h,a)$ for some $a \in \Acal$. That is, $\Ecal_h(f,f')$ denotes the event that $f$ and $f'$ do not fully agree on the safety of $s_h$. Then, for any $\pi \in \Pi$, $h \in [H]$, and $r \geq 0$, we define the pseudometric $\rho_{h,\pi}$ on $\Fcal$ and the corresponding ball $B_{h,\pi}(r)$ about $f^\star$ by
\begin{align*}
    \rho_{h,\pi}(f,f') &= \PP_\pi(\Ecal_h(f,f')) \\
    B_{h,\pi}(r) &= \{ f \in \Fcal : \rho_{h,\pi}(f,f^\star) \leq r\} \,.
\end{align*}
Then, for any $\pi \in \Pi$, $h \in [H]$, and $r_0 \geq 0$, we define the disagreement coefficient $\theta_{h,\pi}(r_0)$ by
\begin{equation*}
    \theta_{h,\pi}(r_0) = \sup_{r > r_0} \frac{\PP_\pi(\exists f,f' \in B_{h,\pi}(r) : \Ecal_h(f,f'))}{r} \,.
\end{equation*}
Note that clearly by construction $\theta_{h,\pi}(r_0)$ is non-increasing in $r_0$. Then, we require the following technical assumption.
\begin{assumption}
\label{assum:theta}
    There exists some fixed $\thetamax < \infty$ such that $\theta_{h,\pi}(0) \leq \thetamax$ for all $h \in [H]$ and $\pi \in \pisafe$.
\end{assumption}
Note that for some problems we may have $\theta_{h,\pi}(0) = \infty$. In this case, we can still obtain a similar PAC bound given a complex technical condition in terms of the rate of growth of $\theta_{h,\pi}(r)$ as $r \to 0$, which we present in the appendix. However, we focus here on using \pref{assum:theta} instead since it is much simpler and already covers many important settings, such as linear classifiers with bounded density \citep{hanneke2014theory}. We also refer readers to \citet{hanneke2014theory} for a detailed discussion of known results on disagreement coefficients.

Given these assumptions, we are ready to present our main theoretical result.
\begin{theorem}
\label{thm:main}
    Let $\vcf$ denote the VC dimension of $\Fcal$, and let some $\epsilon,\delta \in (0,1)$ be given. Suppose we run \algname with $N=H$, $B = n_B(\epsilon,H,\pcd)$, $m=n_m(\epsilon,\delta,H,\pcd,\thetamax,\vcf)$, $\epsexplore = \frac{1}{8} H^{-2} \epsilon$, $\epsfinal = \frac{1}{2} \epsilon$, $\deltaexplore = \frac{1}{4NB} \delta$, and $\deltafinal = \frac{1}{2} \delta$, for some
    \begin{align*}
        n_B &= \Ocal \Big( \log(\epsilon^{-1}) \log(H) \pcd \Big)  \\
        n_m &= \Ocal \Big( \epsilon^{-1} \log(\delta^{-1}) H^3 \log(H)^2 \thetamax^2 \vcf \Big) \,.
    \end{align*}
    Then, under \Cref{assum:blackbox-rl,assum:safety-realizability,assum:policy-cover,assum:theta}, the returned policy $\hat\pi$ satisfies $\SubOpt(\hat\pi;\pisafe) \leq \epsilon$ with probability at least $1-\delta$.
    
    Ignoring log terms, it follows that the total sample complexity (number of episodes) of our algorithm is at most
    \begin{align*}
        n_{\textup{sample}} = \widetilde\Ocal&\Big( H^4 \epsilon^{-1} \log(\delta^{-1}) \pcd \thetamax^2 \vcf \\
        &\quad + H \pcd n_{\Alg}(H^{-2} \epsilon, \delta) \Big) .
    \end{align*}
    Finally, under an additional event with probability at least $1-\delta$, the number of calls to the labeling oracle with the above settings is bounded by 
    \begin{equation*}
        n_{\textup{label}} = \widetilde\Ocal \Big( A \log(\epsilon^{-1}) \log(\delta^{-1}) H^2  \pcd \thetamax^2 \vcf \Big) \,.
    \end{equation*}
\end{theorem}

Importantly, \pref{thm:main} tells us that although we require a sample complexity that depends on $\epsilon$ and $H$ by at least $\widetilde\Ocal(\epsilon^{-1} H^4)$ (possibly greater depending on the complexity of the blackbox algorithm), the corresponding dependence for the sampling oracle is \emph{much} smaller at $\widetilde\Ocal(\log(\epsilon^{-1}) H^2)$. This is very desirable compared with naive baselines, which may require a number of labels on the same order as the sample complexity. We also note that finite classes $\Fcal$ have VC dimension at most $\log_2(|\Fcal|)$, so in the case that $\Fcal$ is finite we can replace $\vcf$ by $\log(|\Fcal|)$.

\paragraph{Proof Overview}
Here we provide a brief overview of the proof of \pref{thm:main}. Full proof details, along with a generalized result using a relaxed version of \pref{assum:theta}, as discussed above, are provided in the supplement.

First, by \pref{lem:subopt-bound} and the guarantees of the blackbox algorithm, if we can ensure $U(\Dcal_N^{(B)}) \leq \frac{1}{2} H^{-1} \epsilon$ with probability at least $1-\frac{1}{2} \delta$, then we have $\SubOpt(\hat\pi;\pisafe) \leq \epsilon$ with probability at least $1-\delta$ by a union bound. Therefore, the proof focuses on establishing this bound.

Now, define $G(\pi;\Dcal) = \frac{1}{H} \sum_{h=1}^H \PP_\pi(\exists a : s_h \in \rd^a(\Dcal))$,
and $\Delta = \frac{1}{4} H^{-3} \epsilon$.
Given \pref{assum:theta}, we show that our choice of $m$ ensures that, after rolling out with any fixed $\pi$ for $m$ iterations to expand $\Dcal$ to $\Dcal'$, we will have $G(\pi;\Dcal') \leq \frac{1}{2} \max(\Delta, G(\pi;\Dcal))$ with high probability.

Furthermore, given \pref{assum:policy-cover} and any fixed $\Pi$, if we find $\hat\pi$ such that $G(\hat\pi;\Dcal) \geq \sup_{\pi \in \Pi} G(\pi;\Dcal) - \frac{1}{2} \Delta$ (which is ensured by our choice of $\epsexplore$, since the expected sum of rewards $\tilde R_n^{(i)}$ under $\pi$ is given by $H G(\pi;\Dcal_n^{(i)})$) and expand $\Dcal$ to $\Dcal'$ such that $G(\hat\pi;\Dcal') \leq \frac{1}{2} \max(\Delta, G(\hat\pi;\Dcal))$ at least $B$ times, then at end of the process we have $\sup_{\pi \in \Pi} G(\pi; \Dcal) \leq \Delta$ with certainty. That is, putting the above together, the inner loop of our algorithm ensures with high probability that $\sup_{\pi \in \Pi_n} G(\pi; \Dcal_n^{(B)}) \leq \Delta$.

Finally, we show that after expanding $\Dcal$ to $\Dcal'$ to ensure that $\sup_{\pi \in \Pi(\Dcal)} G(\pi; \Dcal') \leq \Delta$ at least $H$ times, we are ensured with certainty that at the end of this process we have $U(\Dcal) \leq \frac{1}{2} H^{-1} \epsilon$. This follows, intuitively, because after repeating the above process $h$ times, we will always know safety values for the first $h$ actions of any safe policy.

\subsection{Examples}
\label{sec:examples}

\begin{table*}[t]
    \centering
    {\renewcommand{\arraystretch}{1.25} 
    \begin{tabular}{ccccc}
        \hline
        $\Mcal$ & $\pcd$ bound & $\Alg$ & $n_{\Alg}$ & \algname sample complexity  \\
        \hline
        Tabular MDP & $S$ & UCB-VI & $\widetilde\Ocal(H^3 S^3 A \epsilon^{-2} \log(\delta^{-1})^4)$ & $\widetilde\Ocal(H^8 S^4 A \epsilon^{-2} \log(\delta^{-1})^4 \thetamax^2 \vcf )$ \\
        Block MDP & $S$ & HOMER & $\widetilde\Ocal(H^4 S^8 A^4 \epsilon^{-2} \log(\delta^{-1}) )$ & $\widetilde\Ocal(H^9 S^9 A^4 \epsilon^{-2} \log(\delta^{-1}) \thetamax^2 \vcf )$ \\
        Low NNR MDP & $\dnnr$ & Rep-UCB & $\widetilde\Ocal(H^5 d^4 A^2 \epsilon^{-2} \log(\delta^{-1}) )$ & $\widetilde\Ocal(H^{10} d^5 A^2 \epsilon^{-2} \log(\delta^{-1}) \thetamax^2 \vcf )$ \\
        \hline
    \end{tabular}
    }
    \caption{Summary of example instantiations of our theory for different MDP classes $\Mcal$, and corresponding blackbox algorithms $\Alg$. We consider UCB-VI \citep{azar2017minimax}, HOMER \citep{misra2020kinematic}, and Rep-UCB \citep{uehara2021representation}. In each case we give a bounds $\pcd$, the $n_{\Alg}$, and the corresponding sample complexity of \algname.}
    \label{tab:examples}
\end{table*}

Finally we provide some examples that instantiate our theory to some particular classes of MDPs. For each MDP class $\Mcal$ considered below we will provide a bound on $\pcd$, an example blackbox algorithm $\Alg$ for this class, its sample complexity $n_\Alg$, and the corresponding sample complexity of \algname. We provide all of these bounds in \pref{tab:examples}, and provide details on how the bounds on $\pcd$ are derived in the appendix.

\paragraph{Tabular MDPs}
Tabular MDPs is a simple MDP class where $\Scal$ is finite, and no other structure is assumed. Letting $|\Scal| = S$, it is easy to show that $\pcd \leq S$. In this example, we consider the UCB-VI algorithm \citep{azar2017minimax}, which is based on value iteration with an exploration reward bonus, and is known to be minimax optimal for this class. Note that \citet{azar2017minimax} only provided a regret bound; we provide details of how we converted this to a corresponding sample-complexity bound in the appendix.

\paragraph{Block MDP}
Block MDP is an MDP class where $\Scal$ can be general, but the transition dynamics are defined by an latent tabular MDP. Specifically, the observed states are sampled iid given the latent discrete state, and it is assumed that the observed state distributions for each latent state do not overlap.\footnote{This ensures that the latent states are fully recoverable from the observed states, and therefore the process is an MDP.} Let $S$ denote the number of discrete latent states. Then, it can be shown that we again have $\pcd \leq S$. Here we consider the HOMER algorithm of \citet{misra2020kinematic}, which works by systematically exploring the latent state space, while simultaneously learning a decoder to map observed to latent states.

\paragraph{Low NNR MDP}
Finally, we consider the class of MDPs with low non-negative rank (NNR). We define the NNR of an MDP as the smallest integer $\dnnr$ such that the transition operator can be written as $T(s' \mid s, a) = \sum_{z=1}^{\dnnr} \PP(z \mid s, a) \PP(s' \mid z)$, for some latent variable $z \in [\dnnr]$. It can easily be shown that this class generalizes both Block MDP and Tabular MDP, and that we always have $\pcd \leq \dnnr$. For this example we consider the Rep-UCB algorithm \citep{uehara2021representation}, which is an algorithm with an optimism-based exploration bonus, which simultaneously learns the low-rank decomposition while exploring.

\section{Related Work}
\label{sec:related-work}

There has been a vast body of work on safe RL in recent years; for an overview see for example \citet{garcia2015comprehensive}, \citet{gu2022review}, or \citet{brunke2022safe}. Past work has considered many different approaches, including heuristic deep reinforcement learning methods \citep{thomas2021safe,luo2021learning}, providing PAC bounds on both sample complexity and number of safety violations \citep{ding2021provably,hasanzadezonuzy2021learning}, safety in the context of transfer learning \citep{srinivasan2020learning}, safe offline RL \citep{amani2022doubly}, or safe multi-agent RL \citep{lu2021decentralized}.

Within the vast body of work on safe RL, there is a sub-area of particular relevance that focuses on guaranteeing safety during training. One significant line of work here focuses on the setting where the safety function is unknown and numeric safety feedback is given, and work by leveraging smoothness assumptions on the safety function \citep{sui2015safe,turchetta2016safe,wachi2018safe,wachi2020safe,cheng2019end}. However, these papers require numeric feedback and deterministic state transitions, and are limited to continuous control-like settings.
A different line focuses on settings where the safety function is known, with diverse approaches including safe versions of value and policy iteration \citep{chow2018lyapunov}, reducing to blackbox optimal control problems given differentiability of trajectories \citep{jin2021safe}, or using MDP abstractions \citep{simao2021alwayssafe}.
However, these works are not relevant when safety must be learned. Other works in this sub-area include methods for low-rank MDPs with linear safety functions and numeric feedback \citep{amani2021safe}, settings with cost-based safety functions given a total safety budget \citep{huang2022safe}, or using a human-in-the-loop who can take control in real time \citep{saunders2018trial}. However, these approaches all fail to meet our requirement of being able to provably ensure safety during training given offline binary feedback. Alternatively, \citet{roderick2021provably} using ``analogy'' relationships to expand known sets of safe state, action pairs, but this is limited to settings where such a relationship is available. We also note that \emph{none} of the existing papers in this sub-area consider the problem of actively acquiring safety feedback, or minimizing the amount of safety feedback required.

\begin{figure*}
    \centering
    \includegraphics[scale=0.33]{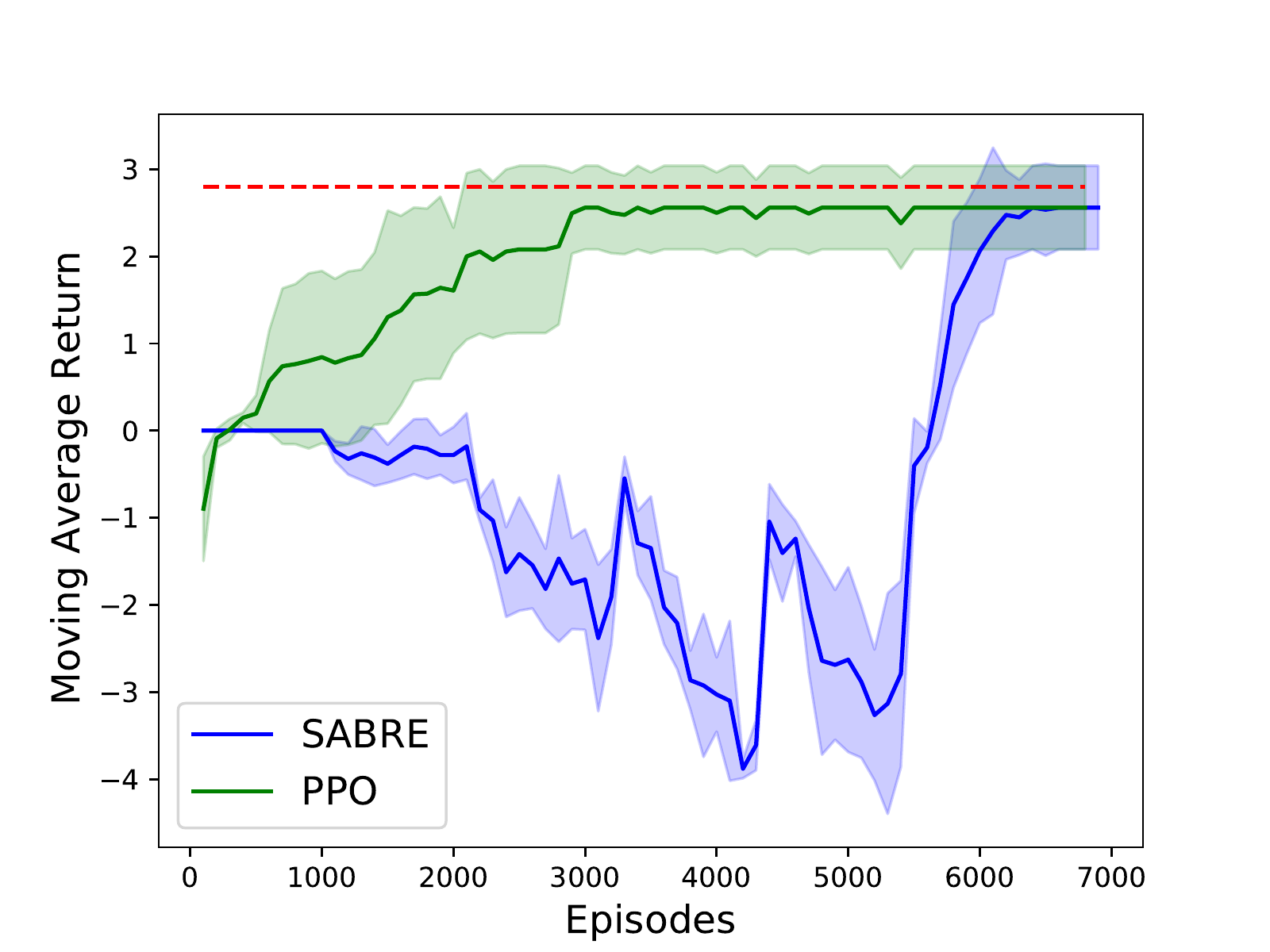}
    \includegraphics[scale=0.33]{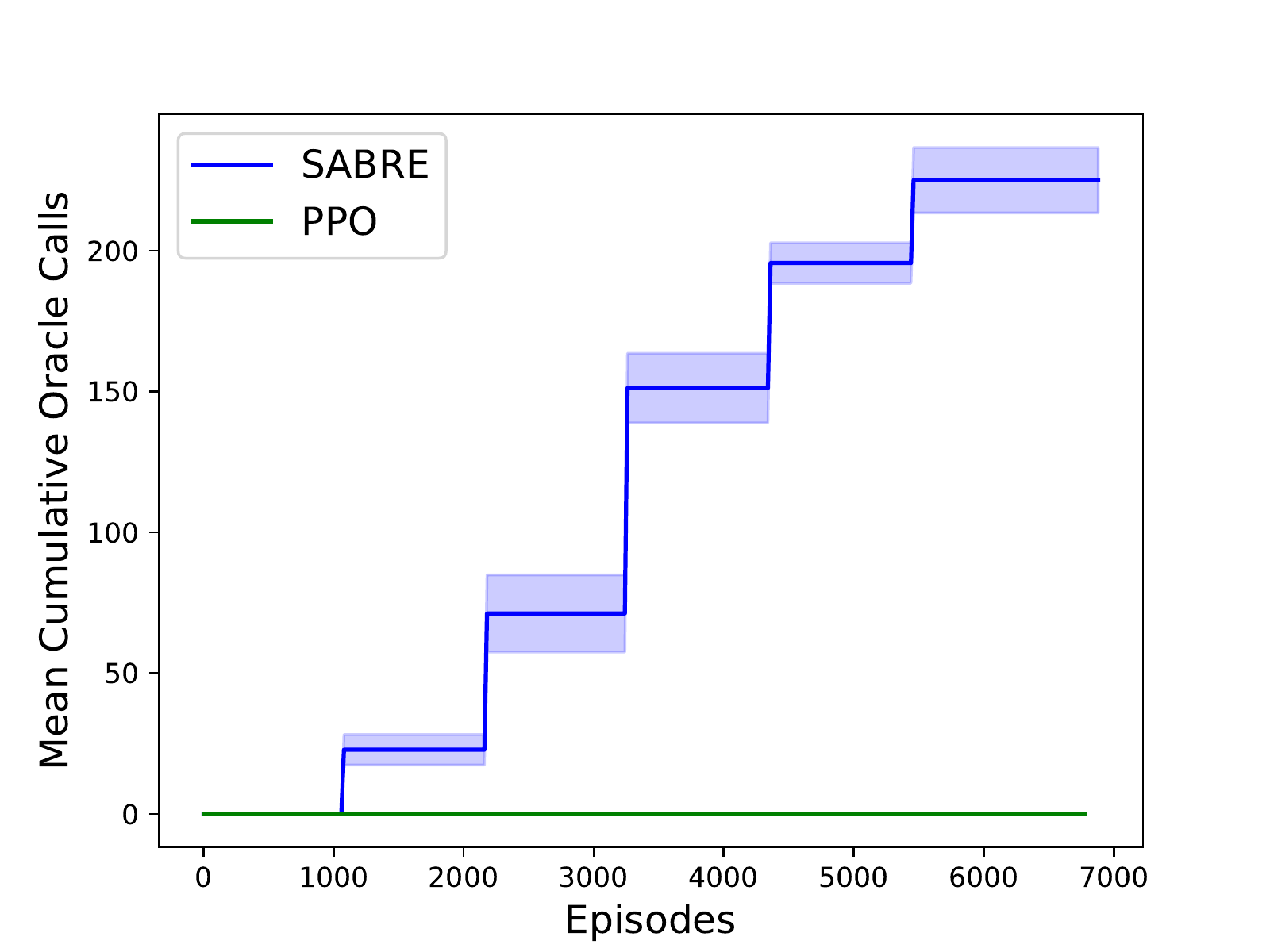}
    \includegraphics[scale=0.33]{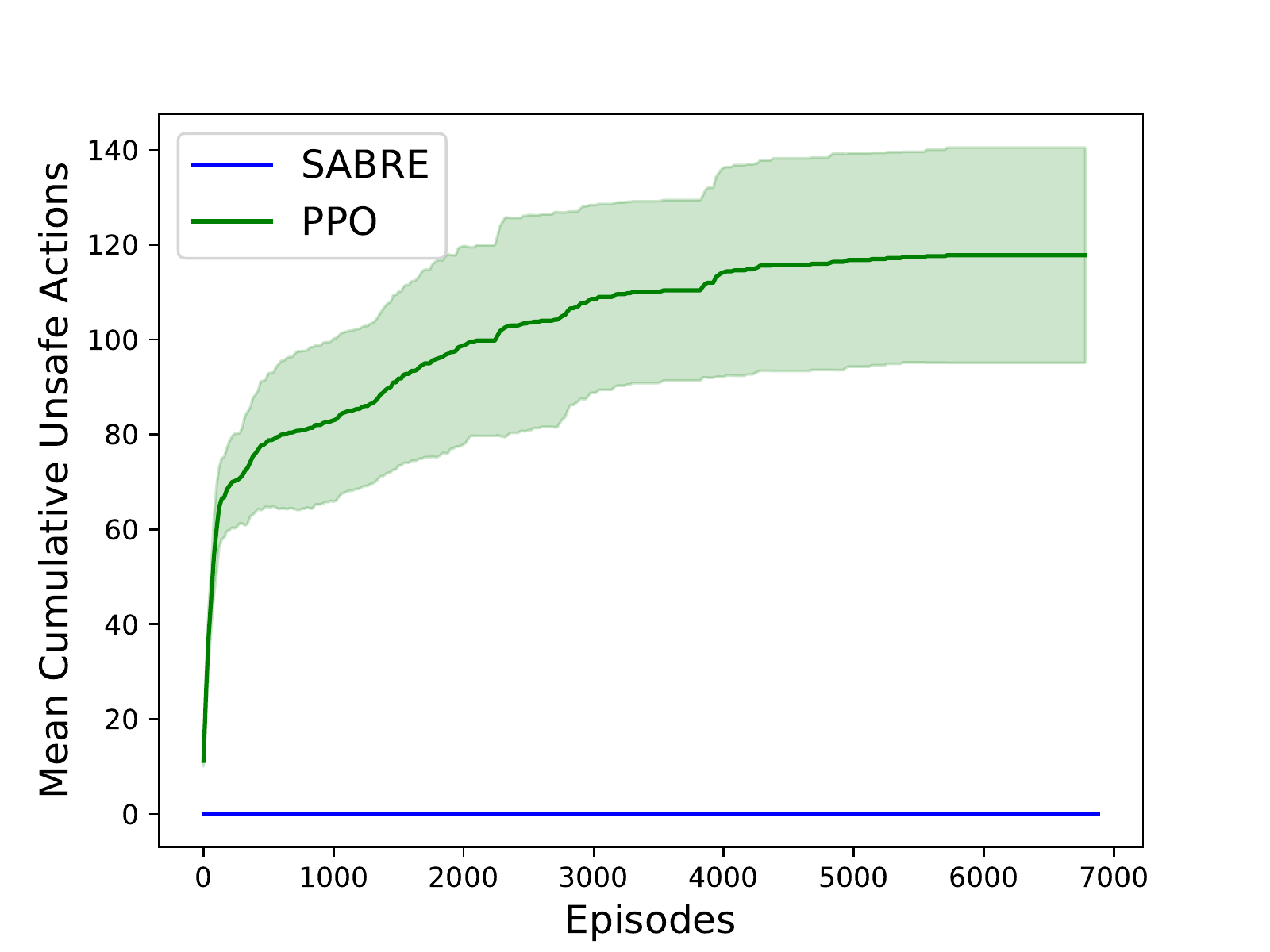}
    \caption{\textbf{Left:} Mean return of $\algname$ on the block MDP task. The red line shows the optimal return of $2.8$. \textbf{Middle:} Mean cumulative number of calls to the labeling oracle over episodes. \textbf{Right:} Mean cumulative number of unsafe actions over episodes. For all plots the error bars correspond to one standard deviation of the mean statistic (estimated over the five replications).}
    \label{fig:results}
\end{figure*}

Finally, another relevant set of literature is on disagreement-based active learning in the realizable setting \citep{hanneke2014theory}.
In particular, both our algorithm and analysis adopt ideas from the CAL algorithm of \citet{cohn1994improving}, which is a simple approach that labels whenever we observe an input in the region of disagreement, and enjoys both sample- and label-complexity bounds in terms of the disagreement coefficient \citep{hanneke2014theory}.
Other works related to this consider, for example, bounding disagreement coefficients for particular settings (for a detailed overview see \citet{hanneke2014theory} and references therein), or providing lower and upper bounds for the problem of realizable active learning \citep{dasgupta2005coarse,balcan2007margin}. However, these works all consider standard binary classification settings where data is sampled from a fixed distribution, rather than RL settings where we would like to learn classifiers (for safety) that are accurate under the state distributions induced by a wide range of policies.

\section{Proof of Concept Experiments}

Finally, we provide some brief ``proof of concept'' experiments to help highlight the correctness of our theory. The focus here is to demonstrate that \algname can achieve near-optimal return in a reasonable number of episodes in a non-trivial scenario, while never taking unsafe actions.

\paragraph{Environment}
We consider a particular Block MDP scenario, which was introduced in \pref{sec:examples}. This scenario has $A=4$ actions, a time horizon of $H=5$, and $S=4H+1$ discrete latent states. The agent receives an observed state $s$ along with a safety feature $\phi(s, a)$ for every action $a$. The underlying latent state space contains four different paths the agent may take: an \emph{unsafe} path that the agent follows by taking unsafe actions; a \emph{high reward} path where the agent receives high rewards but learns the safety features more slowly; a \emph{low reward} path where the agent receives low rewards but learns safety features more quickly; and a \emph{safe path} which the agent reaches by following $\piknown$ which is absorbing and gives no reward. The idea of the scenario is that, to safely optimize reward quickly, the agent should first follow the low-reward path to learn the safety function quickly, rather than greedily try to follow the high-reward path where the safety function is learned more slowly. We provide full details of this environment in the supplement.

\paragraph{Blackbox Algorithm} We use Proximal Policy Optimization as our blackbox RL algorithm~\cite{schulman2017proximal}.  PPO is a popular empirical RL method and while not a provably efficient algorithm, is quite effective in practice for problems that do not require strategic exploration. We describe the specifics of our PPO implementation in the supplement.

\paragraph{Safety Class} We let the safety class be given by $\Fcal = \{ \sign(w^\top \phi(s,a)): \|w\|_\infty \leq 1\}$. We note that checking whether $s \in \rd^a(\Dcal)$ for any $(s,a)$, as required by $\algname$, can be reduced to solving linear programs. We provide details of this in the supplement.

\paragraph{Results} 
We ran \algname, implemented with the PPO blackbox algorithm as discussed above on our Block MDP environment, over a total of 7000 episodes. In addition, for comparison we ran the PPO algorithm, directly optimizing reward without any safety considerations. We repeated this over 5 random replications, and present results in~\pref{fig:results}.

The left-most figure shows the average episodic return against episodes for both \algname and PPO.
In the case of \algname, as expected, the agent initially receives negative reward as it explores the safety relation along the low reward path. Then, near the end, once the safety relation is approximately known, it computes a near-optimal policy. On the other hand, PPO takes a more direct path, although they both end up with a similarly optimal policy. 

The middle figure shows the median number of calls to the labeling oracle over the same set of episodes. The total number corresponds to only approximately $0.2\%$ of the total number of state, action pairs encountered. 
Finally, on the right we plot the average number of safety violations of the two methods. As expected, the unsafe PPO algorithm takes many unsafe actions. On the other hand, \algname never takes any unsafe actions, as guaranteed by our theory.

In summary, $\algname$ is able to achieve optimal return similar to a standard PPO baseline, while taking no unsafe actions,
and only making a tiny number of oracle calls.
This supports the theoretical findings described above.

\section{Conclusion}

We presented a novel algorithm, \algname, which addresses the problem of safe RL, where the safety function must be learned via binary feedback that is actively acquired in between episodes. Under appropriate assumptions, \algname is guaranteed to return a policy that only takes safe actions, and to never take unsafe actions during training. In addition,
given access to a PAC blackbox RL algorithm for optimizing arbitrary reward functions in the underlying MDP class, it is guaranteed with high probability an approximately-optimal safe policy with polynomial sample complexity. Furthermore, it only requires labels for a relatively minimal number of state, action pairs, with a label complexity that scales in $H^2 \log(\epsilon^{-1})$ as opposed to a sample complexity that scales in $H^4 \epsilon^{-1}$.

There are multiple avenues for future research. First, improved concepts from realizable active learning could be used to further reduce the label complexity of the algorithm, by using a more sophisticated strategy rather than always labeling states in the region of disagreement \citep{dasgupta2005coarse}. Second, it is possible that an improved theoretical analysis could find settings under which the simpler baseline approach described in \pref{sec:baseline-approach} has some provable guaranteed. Similarly, an improved theoretical analysis could allow \algname to be simplified, and only need a single loop where the policy class is updated every iteration. Fuure work may also consider how to relax the realizability assumption, and provide some kind of guarantees when function approximation is used to model $f^\star$. This is important, since a major limitation of our work is that the safety guarantees only hold as long as we satisfy \pref{assum:safety-realizability}. Finally, it would be interesting to see applications of our theory to practical safety problems.

\newpage

\bibliography{ref}
\bibliographystyle{abbrvnat}

\newpage
\appendix

\section{General Implementation Concerns for Blackbox RL Algorithm}

One concern an astute reader may have about our requirement of the blackbox algorithm in \pref{assum:blackbox-rl} is that it not only needs to be able to optimize any reward function, but it needs to be able to do so over any given policy set $\Pi'$, while only taking actions in $\Pi'$. While this may seem like a major requirement and limitation, we explain here why this is not so, and provide a general recipe for dealing with these constraints.

First, note that the policy sets $\Pi'$ that we require $\Alg$ to be actually able to work with are all of the form $\Pi(\Dcal)$. These all look like $\Pi$, except with some constraints added on which actions are allowed in any given state (\emph{i.e.} based on whether or not that state, action pair is known to be safe or not given $\Dcal$). Now, for any given $\Dcal$, let us consider the MDP $M(\Dcal)$, which is defined identically to the actual MDP $M$, except whenever the agent would take an action $a$ in state $s$ that is not known to be safe according to $\Dcal$ the MDP transitions following $T(\cdot \mid s, \piknown(s))$ rather than $T(\cdot \mid s, a)$. Next, note that if we add a layer of abstraction between the agent and the MDP $M$, which checks any action the agent takes, and converts it to $\piknown(s)$ if it is not known to be safe, then interacting with $M$ via this abstraction layer is equivalent to interacting with $M(\Dcal)$. 

Now, let $\pi^\star$ be an $\epsilon$-optimal policy over $\Pi$ for $M(\Dcal)$, and $\tilde\pi^\star$ be the policy given by following $\pi^\star$, and converting its action to $\piknown(s)$ whenever the action is not known to be safe given $\Dcal$.
By construction, following $\pi^\star$ in $M(\Dcal)$ is equivalent to following $\tilde\pi^\star$ in $M$, and also by construction the optimal return in $M$ over $\Pi(\Dcal)$ is the same as the optimal return in $M(\Dcal)$ over $\Pi$.
Therefore, it is clear that we have $\SubOpt(\tilde\pi^\star; \Pi(\Dcal)) \leq \epsilon$. Furthermore, as long as the class $\Pi$ is not defined in some absurd or pathological way, we should have $\tilde\pi^\star \in \Pi$, in which case we would also have $\tilde\pi^\star \in \Pi(\Dcal)$. That is, $\tilde\pi^\star$ would be an $\epsilon$-optimal policy over $\Pi(\Dcal)$ for $M$. Note that the above condition that $\tilde\pi^\star \in \Pi$ is trivial for example if $\Pi$ consists of all policies, or all deterministic policies.

Given the above reasoning, as long as the MDP $M(\Dcal)$ is still within the class $\Mcal$, we can implement the blackbox algorithm by estimating an $\epsilon$-optimal policy for $M(\Dcal)$ over $\Pi(\Dcal)$, using the kind of abstraction layer discussed above. Since $M(\Dcal) \in \Mcal$, the blackbox PAC guarantees of \pref{assum:blackbox-rl} will still apply to this estimation. The returned policy will then correspond to an $\epsilon$-optimal policy over $\Pi(\Dcal)$, given by converting potentially unsafe actions to $\piknown(s)$ as required.
That being said, this is not necessarily the best approach for implementing $\Alg$, since optimizing over the duplicated $\piknown(s)$ actions may lead to both computational and statistical inefficiencies. More efficient implementations will be dependent on $\Mcal$ and the choice of blackbox algorithm (we discuss a particular case for a better implementation of PPO with arbitrary $\Pi(\Dcal)$ in our experimental details).

Finally, we note that the above argument relied on the assumption that $M(\Dcal)$ was still in the class $\Mcal$, so the PAC guarantees of the algorithm would still apply to $M(\Dcal)$. This is definitely the case for all of the MDP classes considered in \pref{sec:examples}. For example, in the case of low non-negative rank MDPs, the MDP $M(\Dcal)$ will have non-negative rank less than or equal to that of $M$. To see this, note that for any $(s,a)$ that is known to be safe, we still have the decomposition $T(\cdot \mid s,a) = \sum_{z=1}^d \PP(z \mid s, a) \PP'(\cdot \mid z)$, for some $\PP$ and $\PP'$ and for any $(s,a)$ that is not known to be safe we have $T(\cdot \mid s,a) = \sum_{z=1}^d \PP(z \mid s, \piknown(s)) \PP'(\cdot \mid z)$, which is a decomposition of the same rank. Similar logic easily extends to Tabular and Block MDPs.

\section{\pfref{lem:subopt-bound}}

\begin{proof}

First, let some arbitrary $\pi^\star \in \argmax_{\pi \in \pisafe} V(\pi)$ be given, and let us define a policy $\tilde\pi(\Dcal)$ as follows:
\begin{itemize}
    \item If $s_h \notin \rd^a(\Dcal)$ for any $a \in \Acal$, then $\tilde\pi(\Dcal)(\cdot \mid s_h) = \pi^\star(\cdot \mid s_h)$
    \item Otherwise, $\tilde\pi(\Dcal)(\cdot \mid s_h) = \piknown(\cdot \mid s_h)$
\end{itemize}
Note that by construction we have $\tilde\pi(\Dcal) \in \Pi(\Dcal)$, since it only ever takes actions that are known to be safe given $\Dcal$.
In addition, let $\Ecal$ denote the event where $s_h \notin \rd^a(\Dcal)$ for any $a \in \Acal$ or $h \in [H]$. Note that under event $\Ecal$, it follows $\tilde\pi(\Dcal)$ takes actions identically to $\pi^\star$. Therefore, we have
\begin{equation*}
    \SubOpt(\tilde\pi(\Dcal);\pisafe) \leq H \PP_{\tilde\pi(\Dcal)}(\neg \Ecal) \,,
\end{equation*}
since the difference in the sum of rewards received between $\pi^\star$ and $\tilde\pi(\Dcal)$ can never be greater than $H$.
Furthermore, by the definition of $U(\Dcal)$ we have $\PP_\pi(\neg \Ecal) \leq U(\Dcal)$ for every $\pi \in \pisafe$, so therefore
\begin{equation*}
    \SubOpt(\tilde\pi(\Dcal);\pisafe) \leq H U(\Dcal) \,.
\end{equation*}
Then, given this, we can bound
\begin{align*}
    \SubOpt(\hat\pi;\pisafe) &= V(\pi^\star) - V(\hat\pi) \\
    &= V(\tilde\pi(\Dcal)) - V(\hat\pi) + \SubOpt(\tilde\pi(\Dcal);\pisafe) \\
    &\leq \SubOpt(\hat\pi;\Pi(\Dcal)) + \SubOpt(\tilde\pi(\Dcal);\pisafe) \\
    &\leq \SubOpt(\hat\pi;\Pi(\Dcal)) + H U(\Dcal) \,,
\end{align*}
where the second equality follows by adding and subtracting $V(\tilde\pi(\Dcal))$ and plugging in the suboptimality definition, the first inequality follows since $\tilde\pi(\Dcal) \in \Pi(\Dcal)$ by construction, and the second follows from the previous bounds. This is our required bound, so we can conclude.

\end{proof}

\section{\pfref{thm:main}}

Before we present the main proof, we present some additional lemmas from which the proof can be built. In these lemmas, we will define
\begin{equation*}
    \rd(\Dcal) = \bigcup_{a \in \Acal} \rd^a(\Dcal) \,,
\end{equation*}
for every safety-labeled dataset $\Dcal$. Also, let
\begin{equation*}
    G(\pi;\Dcal) = \frac{1}{H} \sum_{h=1}^H \PP_\pi(s_h \in \rd(\Dcal)) \,,
\end{equation*}
for any policy $\pi$ and safety-labeled dataset $\Dcal$. 

First, the following lemma provides a guarantee on the behavior of the inner loop of \algname, as long as $m$ is sufficiently large.

\begin{lemma}
\label{lem:inner-loop}
Let \pref{assum:policy-cover} be given. Suppose that for some given $n \in [N]$ in the execution of \pref{alg:ours}, we have the events
\begin{equation}
\label{eq:event-approx-opt}
    G(\hat\pi_n^{(i)};\Dcal_n^{(i-1)}) \geq \sup_{\pi \in \Pi_n} G(\pi;\Dcal_n^{(i-1)}) - \frac{1}{2} \Delta \quad \forall i \in [B] \,,
\end{equation}
and
\begin{equation}
\label{eq:event-progress}
    G(\hat\pi_n^{(i)};\Dcal_n^{(i)}) \leq \frac{1}{2} \max \left\{ \Delta, G(\hat\pi_n^{(i)};\Dcal_n^{(i-1)}) \right\} \quad \forall i \in [B] \,,
\end{equation}
for some $\Delta \in (0,1)$. In addition, suppose that $B \geq 8 \pcd \lceil \log_2(\Delta^{-1}) \rceil$. Then, under these conditions, we are ensured that
\begin{equation*}
    \sup_{\pi \in \Pi_n} G(\pi;\Dcal_n^{(B)}) \leq \Delta \,.
\end{equation*}

\end{lemma}

Next, the following lemma analyzes the outer loop of \algname, under the assumptions and events detailed in \pref{lem:inner-loop}.

\begin{lemma}
\label{lem:outer-loop}

Suppose that the conditions of \pref{lem:inner-loop} holds for every epoch $n \in [N]$, with $\Delta \leq \frac{1}{2} H^{-2} \epsilon$ for some $\epsilon \in (0,\frac{1}{2})$. Suppose also, that $N \geq H$. Then, the final dataset $\Dcal_N^{(B)}$ satisfies
\begin{equation*}
    U(\Dcal_N^{(B)}) \leq \epsilon \,.
\end{equation*}

\end{lemma}

Next, we note that in order to apply the above lemmas, we need to be able to establish the conditions of \pref{lem:inner-loop}. For \pref{eq:event-approx-opt}, we can rely on the guarantees of the blackbox algorithm \Alg. However, for \pref{eq:event-progress}, we can instead appeal to an additional lemma. For this lemma, we introduce the following assumption, which is a weaker version of \pref{assum:theta}.
\begin{assumption}
\label{assum:theta-2}
    There exists some continuous, non-increasing function $\theta^\star : \RR^+ \to \RR^+$ such that
    \begin{equation*}
        \theta^\star(r) \geq \theta_{h,\pi}(r) \quad \forall r > 0, h \in [H], \pi \in \pisafe \,.
    \end{equation*}
    Furthermore, this function satisfies $\lim_{r \to \infty} \theta^\star(r) r = \infty$, and $\lim_{r \to 0} r \theta^\star r = 0$
\end{assumption}

Furthermore, we will define the function $\thetamin : \RR^+ \to \RR^+$, according to
\begin{equation*}
    \thetamin(x) \defeq \inf_{r > 0: r \theta^\star(r) \geq x} \theta^\star(r) \,.
\end{equation*}
We note that given \pref{assum:theta-2}, $\{r > 0 : r \theta^\star(r) \leq x\}$ must be non-empty for every $x>0$, so this is a well-defined function. Also, we note that in the case that $\theta^\star(0) < \infty$, we clearly have $\thetamin(x) \leq \theta^\star(0)$ for every $x>0$, since $\theta^\star$ is non-increasing.

\begin{lemma}
\label{lem:active-learning-bound}
Let \pref{assum:theta} be given. Consider some arbitrary $n \in [N]$ and $i \in [B]$ in the running of \pref{alg:ours}. Then, as long as $m \geq n_m(\Delta,\thetamin,\vcf,\delta,H)$, for some
\begin{equation*}
    n_m(\Delta,\thetamin,\vcf,\delta,H) = \Ocal \Bigg( \Delta^{-1} \thetamin(\Delta/32)^2 \vcf \log(1/\delta) \log(H) \Bigg) \,,
\end{equation*}
we have
\begin{equation*}
    G(\hat\pi_n^{(i)};\Dcal_n^{(i)}) \leq \frac{1}{2} \max \left\{ \Delta, G(\hat\pi_n^{(i)};\Dcal_n^{(i-1)}) \right\} \,,
\end{equation*}
with probability at least $1-\delta$.

\end{lemma}

Given the above lemmas, we can provide the main proof.
\begin{proof}[\pfref{thm:main}]

First, we note that applying \pref{lem:active-learning-bound} with $\delta=\deltaexplore$ and $\Delta=\frac{1}{4} H^{-3} \epsilon$, the corresponding choice of $m$ given by this lemma ensures that \pref{lem:active-learning-bound} holds at each iteration of the algorithm with probability at least $1-\deltaexplore$.
Then, applying a union bound over these $NB$ events, we have that \pref{eq:event-progress} holds for \emph{every} $n \in [N]$ with probability at least $1-\delta/4$.

Next, given the guarantee of \Alg, and the setting of $\epsexplore$ and noting that $H G(\pi;\Dcal_n^{(i-1)})$ corresponds to the expected sum of rewards policy $\pi$ under $\tilde R_n^{(i)}$,  we are ensured with probability at least $1-\deltaexplore$
\begin{align*}
    &H G(\hat\pi_n^{(i)};\Dcal_n^{(i-1)}) \geq \sup_{\pi \in \Pi_n} H G(\pi;\Dcal_n^{(i-1)}) - \frac{1}{8} H^{-2} \epsilon \\
    &\iff G(\hat\pi_n^{(i)};\Dcal_n^{(i-1)}) \geq \sup_{\pi \in \Pi_n} G(\pi;\Dcal_n^{(i-1)}) - \frac{1}{2} \Delta \,,
\end{align*}
at any given $i \in [B]$ and $n \in [N]$. Then, by a union bound over all $NB$ calls to \Alg during the exploratory loops of the algorithm, and noting the value of $\deltaexplore$, we have that \pref{eq:event-approx-opt} holds for every $n \in [N]$ and $\Delta=\frac{1}{4} H^{-3} \epsilon$ with probability at least $1-\delta/4$.
Then, as long as $B$ is at least $8 \pcd \lceil \log_2(\frac{1}{4} H^3 \epsilon^{-1}) \rceil$, the conditions of \pref{lem:inner-loop} hold for every $n \in [N]$, so by \pref{lem:outer-loop} we have
\begin{equation*}
    U(\Dcal_N^{(B)}) \leq \frac{1}{2} H^{-1} \epsilon \,,
\end{equation*}
under the above high probability events. Then, by \pref{lem:subopt-bound}, we have
\begin{align*}
    \SubOpt(\hat\pi;\pisafe) &\leq \SubOpt(\hat\pi;\Pi(\Dcal_N^{(B)})) + H U(\Dcal_N^{(B)}) \\
     &\leq \SubOpt(\hat\pi;\Pi(\Dcal_N^{(B)})) + \frac{1}{2} \epsilon \,.
\end{align*}

Furthermore, following the high-probability guarantee of the final call to \Alg, and the settings of $\epsfinal$ and $\deltafinal$, we have
\begin{equation*}
    \SubOpt(\hat\pi;\Pi(\Dcal_N^{(B)})) \leq \frac{1}{2} \epsilon \,,
\end{equation*}
with probability at least $1-\delta/2$. Then, combining the above three high probability events, and the previous two bounds, a union bound gives us
\begin{equation*}
    \SubOpt(\hat\pi;\pisafe) \leq \epsilon \,,
\end{equation*}
with probability at least $1-\delta$, as required.

Next, let us analyze the above choices of $m$ and $B$. Plugging in the settings for $N$ and $B$, the above choices of $m$ and $B$ satisfy
\begin{align*}
    m &= \Ocal\Big( \epsilon^{-1} H^3 \log(H) \thetamin(H^{-3} \epsilon/128)^2 \vcf \log((\delta/(4NB))^{-1}) \Big) \\
    &= \Ocal\Big( \epsilon^{-1} \log\log(\epsilon^{-1}) H^3 \log(H)^2 \log\log(H) \log(\pcd) \thetamin(H^{-3} \epsilon/128)^2 \vcf \log(\delta^{-1}) \Big) \\
    &= \Ocal\Big( \epsilon^{-1} \log(\epsilon^{-1}) H^3 \log(H)^3 \log(\pcd) \thetamin(H^{-3} \epsilon/128)^2 \vcf \log(\delta^{-1}) \Big) \\
\end{align*}
and
\begin{equation*}
    B = \Ocal\Big( \log(\epsilon^{-1}) \log(H) \pcd \Big) \,.
\end{equation*}
Then, throwing away log-factors, and noting again that $N=H$, the total number of iterations is given by
\begin{align*}
    n_{\textup{sample}}(\epsilon,\delta,H,\pcd,\thetamax,\vcf) &= NB \Big( n_{\Alg}\Big(\frac{1}{8} H^{-2} \epsilon,\delta/(4NB)\Big) \\
    &\qquad\qquad + \widetilde\Ocal\Big(\epsilon^{-1} H^3 \thetamin(H^{-3} \epsilon / 128)^2 \vcf \log(\delta^{-1})) \Big) \Big) \\
    &= \widetilde\Ocal\Bigg( H \pcd \Big( n_{\Alg}(H^{-2} \epsilon,\delta) + H^3 \thetamin(H^{-3} \epsilon / 128)^2 \vcf \log(\delta^{-1}) \Big) \Bigg) \,,
\end{align*}
where in the second equality we apply the assumption that $n_{\Alg}(\epsilon,\delta)$ is polynomial in $\epsilon^{-1}$ and $\log(\delta)$, so therefore hiding log terms we have $n_{\Alg}(\frac{1}{4} H^{-3} \epsilon,\delta/(4NB)) = \widetilde\Ocal(n_{\Alg}(H^{-3} \epsilon, \delta))$.

Next, let us analyze the number of calls to the labeling oracle. Following the proof of \pref{lem:active-learning-bound}, for any given $\Delta$, after labeling at most $k$ states for each $h \in [H]$, for some
\begin{equation*}
    k = \Ocal(\thetamin(\Delta/32)^2 \vcf \log(\delta^{-1}) \log(H)) \,,
\end{equation*}
we are ensured that, with probability at least $1-\delta$, the probability of encountering additional $s_h \in \rd(\Dcal)$ for each $h \in [H]$ is either reduced by a factor of 4, or that probability was already less than $\Delta/8$. This implies that after at most
\begin{equation*}
   \widetilde\Ocal(H \log(\Delta^{-1}) \thetamin(\Delta/32)^2 \vcf \log(\delta^{-1}) A)
\end{equation*}
queries to the labeling oracle in a given iteration, we are ensured that the probability of encountering $s_h \in \rd(\Dcal)$ is at most $\Ocal(\Delta)$ for every $h \in [H]$, with probability at most $1-\delta$. Then, applying a union bound over the $m$ rollouts in a single iteration with $\Delta = \delta/m$, after the above maximum number of
\begin{equation*}
   \widetilde\Ocal(H \log(m) \thetamin(\delta/(32m))^2 \vcf \log(\delta^{-1})^2 A)
\end{equation*}
queries to the labeling oracle, we will make no further queries in that iteration with probability at least $1-\delta$. Given this, replacing $\delta$ in this bound with $\delta/(2NB)$, and applying a union bound over all iterations, with probability at most $1-\delta$ we make no more than
\begin{align*}
   &\widetilde\Ocal(NB H \log(m) \log(NB) \thetamin(\delta/(64 m N B ))^2 \vcf \log(\delta^{-1})^2 A) \\
   &= \widetilde\Ocal(H^2 \log(\epsilon^{-1}) \pcd \thetamin(\delta/64(NmB))^2 \vcf \log(\delta^{-1})^2 A)
\end{align*}
total calls to the labeling oracle.

Finally, we note that for the simplified version of the results presented in the main paper, we assumed that \pref{assum:theta}, which is stronger than \pref{assum:theta-2}, and implies that we have $\thetamin(x) \leq \thetamax$ for all $x > 0$. Therefore, we can obtain the result in the main paper by replacing $\thetamin(\cdot)$ with $\thetamax$ everywhere.

\end{proof}

\section{Proofs of Additional Lemmas \pref{thm:main}}

\subsection{\pfref{lem:inner-loop}}
\begin{proof}

First, for any $i \in \{0,1,\ldots,B\}$, let
\begin{equation*}
    \mu_i = \sup_{\pi \in \Pi_n} G(\pi;\Dcal_n^{(i)}) \,.
\end{equation*}
Under the events of the lemma statement, we will argue that
\begin{equation}
\label{eq:inner-progress}
    \mu_{8 \pcd +i} \leq \max\Big\{\Delta, \frac{1}{2} \mu_i \Big\} \,,
\end{equation}
for every $i \leq B - 8\pcd$.
Given this result the overall lemma easily follows, by chaining this result $s$ times, where $s = \lceil \log_2(\Delta^{-1}) \rceil$. Under the assumption that $B \geq 8 \pcd s$, and the fact that $\mu_i$ is non-increasing, this gives us $\mu_{B} \leq \mu_{8\pcd s} \leq \max\{\Delta, 2^{-s} \mu_0 \} \leq \Delta$. Note that the last inequality follows since our choice of $s$ satisfies $2^{-s} \mu_0 \leq \Delta$.

Now, let $\tilde\pi_1,\ldots,\tilde\pi_{\pcd} \in \Pi_n$ be the policy cover of $\Pi_n$ ensured by \pref{assum:policy-cover}. Note that for any $\pi \in \Pi_n$ and $i \in [B]$, we have
\begin{align*}
        \sum_{k=1}^{\pcd} \Big( G(\tilde\pi_k;\Dcal_n^{(i-1)}) - G(\tilde\pi_k;\Dcal_n^{(i)}) \Big) &= \sum_{k=1}^{\pcd} \Big( \frac{1}{H} \PP_{\tilde\pi_k}(s_h \in \rd(\Dcal_n^{i-1}) \setminus \rd(\Dcal_n^{i})) \Big) \\
        &\geq \frac{1}{H} \PP_{\pi}(s_h \in \rd(\Dcal_n^{i-1}) \setminus \rd(\Dcal_n^{i})) \\
        &= G(\pi;\Dcal_n^{(i-1)}) - G(\pi;\Dcal_n^{(i)}) \Big) \,,
\end{align*}
where the first and last equalities follow because $\rd(\Dcal_n^{(i)}) \subseteq \rd(\Dcal_n^{(i-1)})$.

Next, let some arbitrary $i \leq B - 8 \pcd$ be given, and suppose that \pref{eq:inner-progress} does not hold. Given this, it must be the case that $\mu_{i+j} > \Delta$ for all $j \in [8 \pcd]$, so for the corresponding iterations \pref{eq:event-progress} can be strengthened to
\begin{equation}
\label{eq:event-progress-2}
    G(\hat\pi_n^{(i+j)};\Dcal_n^{(i+j)}) \leq \frac{1}{2} G(\hat\pi_n^{(i+j)};\Dcal_n^{(i+j-1)}) \quad \forall j \in [8 \pcd] \,.
\end{equation}
Then, for every $j \in [8 \pcd]$ we have
\begin{align*}
    \sum_{k=1}^{\pcd} \Big( G(\tilde\pi_k;\Dcal_n^{(i+j-1)}) - G(\tilde\pi_k;\Dcal_n^{(i+j)}) \Big) &\geq G(\hat\pi_n^{(i+j)};\Dcal_n^{(i+j-1)}) - G(\hat\pi_n^{(i+j)};\Dcal_n^{(i+j)}) \\
    &\geq \frac{1}{2} G(\hat\pi_n^{(i+j)};\Dcal_n^{(i+j-1)}) \\
    &\geq \frac{1}{2} \mu_{i+j-1} - \frac{1}{4} \Delta \\
    &\geq \frac{1}{2} \mu_{i+8\pcd} - \frac{1}{4} \Delta \\
    &> \frac{1}{2} \max\Big\{ \Delta, \frac{1}{2} \mu_i \Big\} - \frac{1}{4} \Delta \\
    &\geq \frac{1}{4} \max\Big\{ \Delta, \frac{1}{2} \mu_i \Big\}  \\
    &\geq \frac{1}{8} \mu_i \,,
\end{align*}
where the strict inequality step follows from the assumption that \pref{eq:inner-progress} doesn't hold. Then, summing across $j$ and noting that this is a telescoping sum, we have
\begin{align*}
    \frac{1}{8} (8 \pcd) \mu_i &< \sum_{k=1}^{\pcd} \Big( G(\tilde\pi_k;\Dcal_n^{(i)}) - G(\tilde\pi_k;\Dcal_n^{(i+8\pcd)}) \Big) \\
    &\leq \sum_{k=1}^{\pcd} G(\tilde\pi_k;\Dcal_n^{(i)}) \\
    &\leq \pcd \sup_{\pi \in \Pi_n} G(\pi;\Dcal_n^{(i)}) \\
    &= \pcd \mu_{i} \,,
\end{align*}
But this is impossible, because we cannot have $\mu_i < \mu_i$. Therefore, we have proved by contradiction that \pref{eq:inner-progress} holds. Also, since we argued the above for an arbitrary $i \leq B - 8 \pcd$, \pref{eq:inner-progress} must hold for \emph{every} such $i$, so we are done.

\end{proof}

\subsection{\pfref{lem:outer-loop}}

\begin{proof}

We will prove that under these conditions, we are ensured that 
\begin{equation}
\label{eq:safety-induction}
    \sup_{\pi \in \pisafe} \PP_{\pi} \Big(\exists h \in [n] : s_h \in \rd(\Dcal_n^{(B)}) \Big) \leq \frac{n}{H} \epsilon  \,,
\end{equation}
for every $n \in [H]$. The required result then follows from this by plugging in $n = H$, since
\begin{equation*}
    U(\Dcal) = \sup_{\pi \in \pisafe} \PP_\pi \Big(\exists h \in [H] : s_h \in \rd(\Dcal) \Big) \,,
\end{equation*}
and because $U(\Dcal_N^{(B)});\pisafe) \leq U(\Dcal_H^{(B)})$, as we assumed $N \geq H$.

In the remainder of this proof, we establish \pref{eq:safety-induction} for every $n \in [H]$ by forward induction on $n$.

\subsubsection*{Base Case}

First, for the base case, we note that after the first inner loop, by \pref{lem:inner-loop} we are ensured that
\begin{equation*}
    \sup_{\pi \in \Pi_1} \frac{1}{H} \sum_{h=1}^H \PP_\pi (s_h \in \rd(\Dcal_1^{(B)})) \leq \frac{1}{2 H^2} \Delta \leq \frac{1}{H^2} \Delta \,.
\end{equation*}
Therefore, by simple algebra, for every $\pi \in \Pi_1$ we have
\begin{equation*}
    \PP_\pi (s_1 \in \rd(\Dcal_1^{(B)})) \leq \frac{1}{H} \Delta \,.
\end{equation*}
Now, for \emph{any} policy $\pi$, the distribution $s_1$ is always the same (it is always equal to the initial state distribution), so therefore the previous inequality is equivalent to
\begin{equation*}
    \sup_{\pi \in \pisafe} \PP_\pi (s_1 \in \rd(\Dcal_1^{(B)})) \leq \frac{1}{H} \Delta \,,
\end{equation*}
which establishes the base case.

\subsubsection*{Inductive Case}

Suppose that \pref{eq:safety-induction} holds for some arbitrary $n-1 \in [H-1]$. We will argue that it then holds for $n$. Now, let
\begin{equation*}
    \pi^\star \in \argmax_{\pi \in \pisafe} \PP_\pi \Big( \exists h \in [n] : s_h \in \rd(\Dcal_{n}^{(B)}) \Big) \,.
\end{equation*}
Also, let $\tilde\pi$ be some element of $\Pi_n$ that is defined identically to $\pi^\star$ at every $s_h$ such that $s_h \notin \rd(\Dcal_n^{(B)})$, and is defined arbitrarily otherwise.
Also, let $E$ denote the event that $s_h \notin \rd(\Dcal_n^{(B)})$ for every $h \in [n-1]$. Note that under event $E$, $\pi^\star$ and $\tilde\pi$ take identical actions at the first $n-1$ time steps, so therefore the distribution of $s_1,\ldots,s_n$ conditional on $E$ is the same under $\pi^\star$ and $\hat \pi$. Therefore, we can bound
\begin{align*}
    &\PP_{\pi^\star} \Big( \exists h \in [n] : s_h \in \rd(\Dcal_{n}^{(B)}) \Big) \\
    &= \PP_{\pi^\star}(E) \PP_{\pi^\star} \Big( \exists h \in [n] : s_h \in \rd(\Dcal_{n}) \mid E\Big) \\
    &\qquad + \PP_{\pi^\star}(\neg E) \PP_{\pi^\star} \Big( \exists h \in [n] : s_h \in \rd(\Dcal_{n}) \mid \neg E\Big) \\
    &\leq \PP_{\pi^\star}(E) \PP_{\pi^\star} \Big( \exists h \in [n] : s_h \in \rd(\Dcal_{n}) \mid E\Big) \\
    &\qquad + \frac{n-1}{H} \Delta \PP_{\pi^\star} \Big( \exists h \in [n] : s_h \in \rd(\Dcal_{n}) \mid \neg E\Big) \\ 
    &\leq \PP_{\tilde\pi} \Big( \exists h \in [n] : s_h \in \rd(\Dcal_{n}) \mid E\Big)  + \frac{n-1}{H} \epsilon \\
    &\leq \frac{\PP_{\tilde\pi} \Big( \exists h \in [n] : s_h \in \rd(\Dcal_{n}) \Big)}{ \PP_{\tilde\pi}\Big(E\Big) }  + \frac{n-1}{H} \epsilon \\
    &\leq \frac{\sum_{h=1}^t \PP_{\tilde\pi} \Big( s_h \in \rd(\Dcal_{n}) \Big) }{ 1 - \frac{n-1}{H} \epsilon }  + \frac{n-1}{H} \epsilon \\
    &\leq 2 \sum_{h=1}^t \PP_{\tilde\pi} \Big( s_h \in \rd(\Dcal_n) \Big)  + \frac{n-1}{H} \epsilon \,.
\end{align*}
Note that in these bounds we apply the inductive assumption, which gives us
\begin{align*}
    \PP_{\pi}\Big( \neg E \Big) &= \PP_\pi \Big( \exists h \in [n-1] : s_h \in \rd(\Dcal_n^{(B)}) \Big) \\
    &\leq \PP_\pi \Big( \exists h \in [n-1] : s_h \in \rd(\Dcal_{n-1}^{(B)}) \Big) \\
    &\leq \frac{n-1}{H} \epsilon \,,
\end{align*}
for every $\pi \in \pisafe$. Also, we apply the assumption that $\epsilon \leq \frac{1}{2}$, which gives us $(1 - \frac{n-1}{H} \epsilon)^{-1} \leq 2$, the fact that $\PP_{\tilde\pi}(\cdot \mid E) = \PP_{\pi^\star}(\cdot \mid E)$, and the fact that $\PP(A \mid B) \leq \PP(A) / \PP(B)$ for generic events $A$ and $B$.

Next, given the assumption that we satisfy the conditions of \pref{lem:inner-loop} with $\Delta=\frac{1}{2}H^{-2} \epsilon$, we have
\begin{align*}
    \sup_{\pi \in \Pi_n} \frac{1}{H} \sum_{h=1}^H \PP_{\pi} \Big( s_h \in \rd(\Dcal_n^{(B)}) \Big) \leq \frac{1}{2 H^2} \epsilon \,.
\end{align*}
This then implies that
\begin{align*}
    \sum_{h=1}^n \PP_{\pi} \Big( s_h \in \rd(\Dcal_n^{(B)}) \Big) \leq \frac{1}{2H} \epsilon \,,
\end{align*}
for every $\pi \in \Pi_n$.

Then, noting that by construction $\tilde\pi \in \Pi_n$, putting together the prior bounds gives us
\begin{align*}
    \PP_{\pi^\star} \Big( \exists h \in [n] : s_h \in \rd(\Dcal_{n}^{(B)}) \Big) &\leq \frac{1}{H} \epsilon + \frac{n-1}{H} \epsilon \\
    &\leq \frac{n}{H} \epsilon \,.
\end{align*}
Finally, noting the definition of $\pi^\star$, the above is equivalent to
\begin{equation*}
    \sup_{\pi \in \pisafe} \PP_{\pi} \Big( \exists h \in [n] : s_h \in \rd(\Dcal_{n}^{(B)}) \Big) \leq \frac{n}{H} \epsilon \,,
\end{equation*}
which completes the inductive case.

\end{proof}

\subsection{\pfref{lem:active-learning-bound}}

\begin{proof}

Let $\Dcal$ be an arbitrary label dataset, $\pi \in \Pi$ be an arbitrary policy to run, and $\Dcal'$ be the dataset that is obtained by running $\pi$ for $m$ episodes, and for all encountered states $s_h$ labeling and adding to $\Dcal$ all $(s_h,a)$ such that $s_h \in \rd^a(\Dcal)$. It is sufficient to argue that, for any choice of $\Dcal$ and $\pi$ such that
\begin{equation}
\label{eq:m-condition}
    G(\pi;\Dcal) \geq \frac{1}{2} \Delta  \,,
\end{equation}
we have with probability at least $1-\delta$ after performing the above process for $m$ episodes that
\begin{equation*}
    G(\pi;\Dcal') \leq \frac{1}{2} G(\pi;\Dcal) \,.
\end{equation*}
In the remainder of the proof, we will consider an arbitrarily chosen $\Dcal$ and $\pi$ that satisfy \pref{eq:m-condition}, and we will reason about how large $m$ needs to be to ensure the above fact.

First, for every $h \in [H]$ and safety labeled dataset $\widetilde\Dcal$, let
\begin{equation*}
    \mu_h(\widetilde\Dcal) = \PP_\pi(s_h \in \rd(\widetilde\Dcal)) \,.
\end{equation*}
Note that according to this definition, we have $G(\pi;\widetilde\Dcal) = \frac{1}{H} \sum_{h=1}^H \mu_h(\widetilde\Dcal)$.
Now, even though \pref{eq:m-condition} ensures that we can bound $\mu_h(\Dcal)$ from below on average across $h \in [H]$, there might be some $h$ for which $\mu_h(\Dcal)$ is arbitrarily small, or even zero. Therefore, it may be very inefficient or even infeasible to try to ensure that $\mu_h(\Dcal') \leq \frac{1}{2} \mu_h(\Dcal)$ simultaneously for every $h \in [H]$.

Instead, we will proceed by defining a target set $\Hcal \defeq \{h \in [H] : \mu_h \geq \frac{1}{8} \Delta\}$ of time indices for which we would like to reduce $\mu_h(\Dcal)$. Suppose we were able to ensure that $\mu_h(\Dcal') \leq \frac{1}{4} \mu_h(\Dcal)$ for all $h \in \Hcal$. Then, we would have
\begin{align*}
    \frac{1}{H} \sum_{h=1}^H \mu_h(\Dcal')  &\leq \frac{1}{4} \frac{1}{H} \sum_{h \in \Hcal} \mu_h(\Dcal) + \frac{1}{H} \sum_{h \notin \Hcal} \mu_h(\Dcal) \\
     &\leq \frac{1}{4} \frac{1}{H} \sum_{h=1}^H \mu_h(\Dcal) + \frac{1}{8} \Delta \\
     &\leq \frac{1}{4} \frac{1}{H} \sum_{h=1}^H \mu_h(\Dcal) + \frac{1}{4}  \frac{1}{H} \sum_{h=1}^H \mu_h(\Dcal) \\
     &= \frac{1}{2} \frac{1}{H} \sum_{h=1}^H \mu_h(\Dcal) \,,
\end{align*}
where in the final inequality we apply \pref{eq:m-condition}. Therefore, if we were able to prove that
\begin{equation}
\label{eq:uh-condition}
    \mu_h(\Dcal') \leq \frac{1}{4} \mu_h(\Dcal) \quad \textup{ w.p. at least } 1-H^{-1} \delta \quad \forall h \in \Hcal \,,
\end{equation}
then our required result would follow by a union bound on $h \in \Hcal$. Therefore, in the remainder of the proof we will consider an arbitrary $h$ such that $h \in \Hcal$.

Next, let $r_h$ defined such that
\begin{equation*}
    r_h \theta^\star(r_h) = \frac{1}{4} \mu_h(\Dcal) \,.
\end{equation*}
Note that by \pref{assum:theta-2}, $r \theta^\star(r)$ is continuous and non-decreasing in $r$ with $\lim_{r \to 0} r \theta^\star(r) = 0$, and $\lim_{r \to \infty} r^\star \theta^\star(r) = \infty$, so clearly such an $r_h$ must exist. Furthermore, given the condition on $\mu_h(\Dcal)$ we have $r_h \theta^\star(r_h) \geq \frac{1}{32} \Delta$, which allows us to bound
\begin{equation*}
    \theta^\star(r_h) \leq \thetamin(\Delta/32) \,.
\end{equation*}

Now, let $\PP^U_{h,\pi}$ denote the distribution of $s_h$ induced by $\pi$, conditional on $s_h \in \rd(\Dcal)$. Given the above definitions, for any given $f \notin B_{h,\pi}(r_h)$, we can bound
\begin{align}
    \PP^U_{h,\pi}(\exists a : f(s_h,a) \neq f^\star(s_h,a)) &= \frac{\PP_{\pi}(\exists a : f(s_h,a) \neq f^\star(s_h,a))}{\mu_h(\Dcal)} \nonumber \\
    &\geq \frac{r_h}{\mu_h(\Dcal)} \nonumber \\
    &= \frac{1}{4} \theta_h(r'_h)^{-1} \nonumber \\
    \label{eq:success-bound}
    &\geq \frac{1}{4} \thetamin(\Delta/32)^{-1} \,.
\end{align}

Next, suppose we draw $k$ iid samples $s_h^{(1)},\ldots,s_h^{(k)}$ from $\PP^U_{h,\pi}$, and let $\Fcal_h = \{f \in \Fcal : f \notin B_{h,\pi}(r_h)\}$. Now, suppose further that it were the case that for every $f \in \Fcal_h$ we had $f(s_h^{(i)},a) \neq f^\star(s_h^{(i)},a)$ for some $i \in [k]$ and $a \in \Acal$. If this were the case, then by labeling these $k$ points we would eliminate all hypotheses outside of $B_{h,\pi}(r_h)$. Therefore, letting $\Dcal_k$ denote the dataset obtained by labeling all of these points and adding them to $\Dcal$, we would have
\begin{align*}
    \mu_h(\Dcal_k) &= \PP_\pi(s_h \in \rd(\Vcal(\Dcal_k))) \\
    &\leq \PP_\pi(s_h \in \rd(B_h(r_h))) \\
    &\leq \theta_{h,\pi}(r_h) r_h \\
    &\leq \theta^\star(r_h) r_h \\
    &= \frac{1}{4} \mu_h(\Dcal) \,,
\end{align*}
which is our required condition. Therefore, we can proceed by: (1) finding $k$ such that the above occurs with probability at least $1-\delta/(2H)$; and (2) finding $m$ such that event $s_h \in \rd(\Dcal)$ occurs at least $k$ times with probability at least $1-\delta/(2H)$. Then, applying a union bound we would be done. We will do these things one at a time.

\subsubsection*{Finding sufficiently large $k$}
First, let us define the following stochastic process indexed by $\Fcal$:
\begin{equation*}
    G_k(f) = \frac{1}{k} \sum_{i=1}^k c(s^{(i)}_h, f) - \EE^U_{\pi,h}[c(s_h, f)] \,,
\end{equation*}
where
\begin{equation*}
    c(s,f) = \one\{\forall a \in \Acal : f(s, a) = f^\star(s, a)\} \,.
\end{equation*}

Then, we have

\begin{align*}
    \sup_{f \in \Fcal_h} \frac{1}{k} \sum_{i=1}^k c(s^{(i)}_h, f) &\leq \sup_{f \in \Fcal_h} \EE^U_{\pi,h}[c(s_h, f)] + \sup_{f \in \Fcal_h} G_k(f) \\
    &\leq 1 - \frac{1}{4} \thetamin(\Delta/32)^{-1} + \sup_{f \in \Fcal_h} G_k(f) \,,
\end{align*}
where in the second inequality follows by applying \pref{eq:success-bound}. Furthermore, we have
\begin{equation*}
    \sup_{f \in \Fcal_h} G_k(f) \leq \Big( \sup_{f \in \Fcal_h} G_k(f) - \EE[\sup_{f \in \Fcal_h} G_k(f)] \Big) + \EE[\sup_{f \in \Fcal_h} G_k(f)] \,,
\end{equation*}
where $\EE[\sup_{f \in \Fcal_h} G_k(f)]$ is the expected value of the random variable $\sup_{f \in \Fcal_h} G_k(f)$ under the $k$ iid draws from $\PP^U_{\pi,h}$ (we will let this distribution be implicit in the rest of this sub-section of the proof). We will proceed by bounding these two terms one by one. Specifically, under any event where both terms are at most $\frac{1}{10} \thetamin(\Delta/32)^{-1}$, we would have $\sup_{f \in \Fcal_k} G_k(f) \leq \frac{1}{5} \thetamin(\Delta/32)^{-1} < \frac{1}{4} \thetamin(\Delta/32)^{-1}$. Therefore, we would have $\sup_{f \in \Fcal_h} \frac{1}{k} \sum_{i=1}^k c(s_h^{(i)}, f) < 1$, which implies that for every $f \in \Fcal_h$ we have $f(s_h^{(i)},a) \neq f^\star(s_h^{(i)},a)$ for some $i \in [k]$ and $a \in \Acal$, which is our required condition.

For the first of these terms, let $\tilde s_h^{(1)}, \ldots, \tilde s_h^{(k)}$ be an arbitrary sequence such that $\tilde s_h^{(i)} = s_h^{(i)}$ for all $i \neq j$, for some $j \in [k]$. Then, we can bound
\begin{align*}
    &\sup_{f \in \Fcal_h} \frac{1}{k} \sum_{i=1}^k c(s^{(i)}_h, f)  - \sup_{f \in \Fcal_h} \frac{1}{k} \sum_{i=1}^k c(\tilde s^{(i)}_h, f) \\
    &\leq \sup_{f \in \Fcal_h} \Big| \frac{1}{k} \sum_{i=1}^k c(s^{(i)}_h, f) - c(\tilde s^{(i)}_h, f)  \Big| \\
    &= \frac{1}{k} \sup_{f \in \Fcal_h} \Big| c(s^{(j)}_h, f) - c(\tilde s^{(j)}_h, f) \Big| \\
    &\leq \frac{1}{k} \,.
\end{align*}
Therefore, we can apply the bounded differences inequality inequality to the first term, which gives us
\begin{equation*}
    \sup_{f \in \Fcal_h} G_k(f) - \EE[\sup_{f \in \Fcal_h} G_k(f)] \leq \frac{1}{10} \thetamin(\Delta/32)^{-1} \,,
\end{equation*}
with probability at least $1 - \exp(-k \thetamin(\Delta/32)^{-2} / 50)$.
We can ensure that this event occurs with probability at least $1-\delta/(2H)$, by noting
\begin{align*}
    &\exp(-k \thetamin(\Delta/32)^{-2} / 50) \leq \delta / (2H) \\
    &\iff k \geq 50 \thetamin(\Delta/32)^2 \log(2H/\delta) \,.
\end{align*}

Next, for the second term above, we will follow a standard symmetrization argument. Let $\epsilon_1,\ldots,\epsilon_k$ be $k$ iid Rademacher random variables (random variables such that $\prob(\epsilon_i=1) = \prob(\epsilon_i=-1) = \frac{1}{2}$ for every $i \in [k]$). Also, let $\bar s_h^{(1)},\ldots,\bar s_h^{(k)}$ be a collection shadow variables that are distributed independent from and identically to $s_h^{(1)},\ldots,s_h^{(k)}$. Then, we have
\begin{align*}
    \EE[\sup_{f \in \Fcal_h} G_k(f)] &= \EE_{s_h^{(1:k)} \sim \PP^U_{\pi,h}}\left[ \sup_{f \in \Fcal_h} \frac{1}{k} \sum_{i=1}^k c(s^{(i)}_h, f) - \EE_{s_h \sim \PP^U_{\pi,h}}[c(s_h, f)] \right] \\
    &= \EE_{\bar s_h^{(1:k)} \sim \PP^U_{\pi,h}}\left[ \sup_{f \in \Fcal_h} \frac{1}{k} \sum_{i=1}^k c(\bar s^{(i)}_h, f) - \EE_{s_h \sim \PP^U_{\pi,h}}[c(s_h, f)] \right] \\
    &\leq \EE_{s_h^{(1:k)}, \bar s_h^{(1:k)} \sim \PP^U_{\pi,h}}\left[ \sup_{f \in \Fcal_h} \frac{1}{k} \sum_{i=1}^k c(\bar s^{(i)}_h, f) - \frac{1}{k} \sum_{i=1}^k c(s_h, f) \right] \\
    &= \EE_{s_h^{(1:k)}, \bar s_h^{(1:k)} \sim \PP^U_{\pi,h}, \epsilon_{1:k} \sim \textup{Unif}(-1,1)} \left[ \sup_{f \in \Fcal_h} \frac{1}{k} \sum_{i=1}^k \epsilon_i \left( c(\bar s^{(i)}_h, f) - c(s_h, f) \right) \right] \\ \\
    &\leq 2 \EE_{s_h^{(1:k)} \sim \PP^U_{\pi,h}, \epsilon_{1:k} \sim \textup{Unif}(-1,1)} \left[ \sup_{f \in \Fcal_h} \frac{1}{k} \sum_{i=1}^k \epsilon_i c(s^{(i)}_h, f) \right] \,,
\end{align*}
where the first inequality follows from Jensen's, the subsequent equality follows by symmetry, and the final inequality follows since $\sup (a + b) \leq \sup a + \sup b$, and since by symmetry on the definitions of $\epsilon_i$, $s_h^{(i)}$, and $\bar s_h^{(i)}$. Next, suppressing the subscript in the expectation for brevity, we note that
\begin{align*}
    2 \EE \left[ \sup_{f \in \Fcal_h} \frac{1}{k} \sum_{i=1}^k \epsilon_i c(s^{(i)}_h, f) \right] &= 2 \EE \left[ \sup_{f \in \Fcal_h} \frac{1}{k} \sum_{i=1}^k \epsilon_i   \one\{\forall a \in \Acal : f(s^{(i)}_h, a) = f^\star(s^{(i)}_h, a)\} \right] \\
    &\leq 2 \EE \left[ \sup_{f \in \Fcal_h} \frac{1}{k} \sum_{i=1}^k \epsilon_i   \one\{f(s^{(i)}_h, a^{(i)}_h) = f^\star(s^{(i)}_h, a^{(i)}_h)\} \right] \\
    &= \EE \left[ \sup_{f \in \Fcal_h} \frac{1}{k} \sum_{i=1}^k \epsilon_i   \left( 2 \one\{f(s^{(i)}_h, a^{(i)}_h) = f^\star(s^{(i)}_h, a^{(i)}_h)\} - 1 \right) \right] \\
    &= \EE \left[ \sup_{f \in \Fcal_h} \frac{1}{k} \sum_{i=1}^k \epsilon_i f(s^{(i)}_h, a^{(i)}_h) \right] \,,
\end{align*}
where the inequality step follows because
\begin{equation*}
    \Big| \one\{\forall a \in \Acal : f(s^{(i)}_h, a)) = f^\star(s^{(i)}_h, a) \} \Big| \leq \EE_{a \sim \PP}\Big[\Big| \one\{f(s^{(i)}_h, a)) = f^\star(s^{(i)}_h, a) \} \Big| \Big] \quad \forall \ \PP \,,
\end{equation*}
and also by applying Jensen's inequality, the second equality follows because $\EE[\sum_{i=1}^k \epsilon_i] = 0$, and the final equality follows because $2 \one\{f(s, a)) = f^\star(s, a)\} - 1 = f(s, a) f^\star(s, a)$, and by symmetry $\epsilon_i f^\star(s^{(i)}_h, a^{(i)}_h)$ and $\epsilon_i$ are identically distributed for each $i \in [k]$.

Next, note that the final bound is the Rademacher complexity of $\Fcal_h$ (under some distribution of state, action pairs). 
Since by assumption $\Fcal$ has VC dimension at most $\vcf$, so does $\Fcal_h$, so by \citet[Theorem 5.6]{rebeschini2020} we have
\begin{equation*}
    \EE \left[ \sup_{f \in \Fcal_h} \frac{1}{k} \sum_{i=1}^k \epsilon_i f(s^{(i)}_h, a)) \right] \leq C \sqrt{\frac{\vcf}{k}} \,,
\end{equation*}
for some universal constant $C$, regardless of the distribution over state, action pairs. Therefore, we have
\begin{equation*}
    \EE\Big[\sup_{f \in \Fcal_h} G_k(f)\Big] \leq C \sqrt{\frac{\vcf}{k}} \,,
\end{equation*}
so we can ensure that $\EE[\sup_{f \in \Fcal_h} G_k(f)] \leq \frac{1}{10} \thetamin(\Delta/32)^{-1}$ as long as
\begin{equation*}
    k \geq 100 C^2 \thetamin(\Delta/32)^2 \vcf \,.
\end{equation*}

Therefore, putting the above together, as long as 
\begin{align*}
    k &\leq  \max\Big(50 \thetamin(\Delta/32)^2 \log(2H / \delta), 100 C^2 \thetamin(\Delta/32)^2 \vcf \Big) \\
    &\leq 50 \thetamin(\Delta/32)^2 (2C^2 \vcf + \log(2H / \delta)) \,,
\end{align*}
then both bounds hold with probability at least $1 - \delta / (2H)$, and therefore we have our required event that every $f \in \Fcal_h$ disagrees with $f^\star$ on at least one of the sampled states $s_h^{(i)}$ with this probability.

\subsubsection*{Finding sufficiently large $m$}

Next, for our target value of $k$ samples from $\PP$, the the remaining question is how large does $m$ need to be to ensure we have at least $k$ samples of $s_h$ where $s_h \in \rd(\Dcal)$, with probability at least $1-\delta/(2H)$? We know that for sampled $s_h$, this event occurs with probability at least $\Delta/8$ by the construction of $\Hcal$. 

Now, suppose we set $m = k (\Delta/8)^{-1} + t$ for some $t \geq 0$, and let $X_1\ldots,X_m$ be a set of iid $\{0,1\}$-valued random variable such that $\prob(X_i=1) = \Delta/8$ for each $i \in [m]$. Also, let $\bar X = \sum_{i=1}^m X_i$. Applying Bernstein's inequality, we have
\begin{align*}
    \prob(\bar X \leq k) &= \prob(\bar X - \EE[\bar X] \leq - t \Delta/8) \\
    &\leq \exp \left( -\frac{\frac{1}{2} t^2 (\Delta/8)^2}{m (\Delta/8)(1 - \Delta/8) + \frac{1}{3} t \Delta/8} \right) \,.
\end{align*}
Now, in order to ensure that this probability is smaller than $\delta/(2H)$, it is sufficient to solve
\begin{align*}
    &\exp \left( -\frac{\frac{1}{2} t^2 (\Delta/8)^2}{m (\Delta/8)(1 - \Delta/8) + \frac{1}{3} t \Delta/8} \right) = \delta/(2H) \\
    &\iff \frac{1}{2} (\Delta/8)^2 t^2 - (4/3 - \Delta/8) (\Delta/8) \log(2H/\delta) t -  (1-\Delta/8) \log(2H/\delta) k = 0 \,.
\end{align*}
Now, letting $t^+$ be the greater solution of this quadratic equation, 
\begin{align*}
    t^\star &= \frac{(4/3 - \Delta/8)(\Delta/8) \log(2H/\delta) }{(\Delta/8)^2} \\
    &\quad + \frac{\sqrt{\Big((4/3 - \Delta/8)(\Delta/8) \log(2H/\delta)\Big)^2 + 2(\Delta/8)^2 (1 - \Delta/8) \log(2H/\delta) k} }{(\Delta/8)^2} \\
    &\leq 2 (4/3 - \Delta/8) (\Delta/8)^{-1} \log(2H/\delta) + (\Delta/8)^{-1} \sqrt{2 (1-\Delta/8) \log(2H/\delta) k} \\
    &\leq 3 (\Delta/8)^{-1} \log(2H/\delta) \sqrt{k} \,,
\end{align*}
where in the first inequality we apply apply the fact that $\sqrt{a+b} \leq \sqrt{a} + \sqrt{b}$ for $a,b > 0$, and in the second inequality we note that $\Delta/8 \leq 1/3$ for any $\Delta \in (0,1)$, so $(4/3-\Delta/8) \leq 1$, and $2(1-\Delta/8) \leq 1$.

Finally, we note that since the probability that $s_h \in \rd(\Dcal)$ is at least $\Delta/8$, the probability that we have fewer than $k$ successes in $m$ samples is no greater than $\prob(\bar X \leq k)$, and therefore from the above bounds any choice of
\begin{equation*}
    m \geq (\Delta/8)^{-1} \Big( k + 3 \log(2H/\delta) \sqrt{k} \Big)
\end{equation*}
is sufficient to ensure that this probability is at most $\delta/(2H)$.

\subsubsection*{Putting Everything Together}

Given the analysis, if we set
\begin{align*}
    k &\geq  50 \thetamin(\Delta/32)^2 \big(2C^2 \vcf + \log(2H / \delta) \big) \\
    \textup{and} \quad m &\geq 8 \Delta^{-1} \Big( k + 3 \log(2H/\delta) \sqrt{k} \Big) \,,
\end{align*}
then our required result is ensured with probability at least $1-\delta$.

We note that this setting of $m$ satisfies the growth bound
\begin{equation*}
    m = \Theta \Bigg( \Delta^{-1} \thetamin(\Delta/32)^2 \vcf \log(1/\delta) \log(H) \Bigg) \,,
\end{equation*}
which is our required result.

\end{proof}

\section{Derivations of Policy Cover Dimension Bounds}

Here we provide the derivations of our policy cover dimension bounds given in \pref{sec:examples}.

\paragraph{Tabular MDP} Let some policy class $\Pi'$ be given, and assume the MDP has $S$ total states, and denote these by $\Scal_1,\ldots,\Scal_S$. Then, for each $i \in [d]$, we define the policy $\pi_i \in \Pi'$ accoding to
\begin{equation*}
    \pi_i = \argmin_{\pi \in \Pi'} \EE_{\pi} \Big[ \sum_{h=1}^H \one\{s_h = \Scal_i\} \Big] \,.
\end{equation*}
That is, $\pi_i$ is the optimal policy in $\Pi'$ for the reward function given by reaching the $i$'th state. Now, let some arbitrary subset $\widetilde S \subseteq S$, and arbitrary $\pi \in \Pi$ be given. Then, we have
\begin{align*}
    \frac{1}{H} \sum_{h=1}^H \PP_\pi(s_h \in \widetilde S) &\leq \frac{1}{H} \sum_{h=1}^H \sum_{s \in \widetilde S} \PP_\pi(s_h = s) \\
    &\leq \frac{1}{H} \sum_{h=1}^H \sum_{i=1}^S \PP_\pi(s_h = \Scal_i) \\
    &= \sum_{i=1}^S \EE_\pi \Big[ \frac{1}{H} \sum_{h=1}^H  \one\{ s_h = \Scal_i \} \Big] \\
    &\leq \sum_{i=1}^S \EE_{\pi_i} \Big[ \frac{1}{H} \sum_{h=1}^H  \one\{ s_h = \Scal_i \} \Big] \,.
\end{align*}
Since the above holds for arbitrary $\widetilde\Scal$ and $\pi$, we have that $\{\pi_1, \ldots, \pi_S\}$ is a policy cover for $\Pi'$. That is, we always have a policy cover of size at most $S$, so $\pcd \leq S$.

\paragraph{Block MDP and Non-negative Rank MDP} The reasoning here in both cases is similar to the tabular MDP case. In both cases, we can model the MDP as transitioning from $(s,a)$ to an intermediate discrete latent state $z$ following some distribution $\PP(\cdot \mid s,a)$, and then sampling the next state $s'$ following another distribution $\PP'(\cdot \mid z)$. Let $d$ be the number of discrete latent states, which corresponds to $S$ for Block MDP and $\dnnr$ for Non-negative Rank MDP in terms of our notation in \pref{sec:examples}. Also, for each $h \in [h]$, let $z_h$ denote the discrete latent state that generated $s_h$ at time $h$, and let $\Zcal_1,\ldots,\Zcal_d$ denote the possible values of $z$. Then, following analogous reasoning as above, define
\begin{equation*}
    \pi_i = \argmin_{\pi \in \Pi'} \EE_{\pi} \Big[ \sum_{h=1}^H \one\{z_h = \Zcal_i\} \Big] \,,
\end{equation*}
for some given policy class $\Pi'$
Also, let some arbitrary measurable $\widetilde\Scal \subseteq \Scal$ be given, along with some arbitrary policy $\pi \in \Pi'$. Then, we have
\begin{align*}
    \frac{1}{H} \sum_{h=1}^H \PP_\pi(s_h \in \widetilde S)  &= \frac{1}{H} \sum_{h=1}^H \sum_{i=1}^d \PP_\pi(z_h = \Zcal_i) \PP'(\widetilde\Scal \mid z = \Zcal_i) \\
    &= \sum_{i=1}^d \PP'(\widetilde\Scal \mid z = \Zcal_i) \EE_\pi \Big[ \frac{1}{H} \sum_{h=1}^H  \one\{ z_h = \Zcal_i \} \Big] \\
    &\leq \sum_{i=1}^d \PP'(\widetilde\Scal \mid z = \Zcal_i) \EE_{\pi_i} \Big[ \frac{1}{H} \sum_{h=1}^H  \one\{ z_h = \Zcal_i \} \Big] \\
    &= \frac{1}{H} \sum_{h=1}^H \sum_{i=1}^d \PP_{\pi_i}(z_h = \Zcal_i) \PP'(\widetilde\Scal \mid z = \Zcal_i) \\
    &\leq \sum_{i=1}^d \frac{1}{H} \sum_{h=1}^H \sum_{j=1}^d \PP_{\pi_i}(z_h = \Zcal_j) \PP'(\widetilde\Scal \mid z = \Zcal_j) \\
    &= \sum_{i=1}^d \frac{1}{H} \sum_{h=1}^H \PP_{\pi_i}(s_h \in \widetilde\Scal) \,,
\end{align*}
where the the above bounds we apply the fact that the distribution $\PP'$ for generating $s$ given $z$ does not depend on either the policy or $h$, and the second inequality follows because we are just introducing additional non-negative summands.
That is, we always have a policy cover of size $d$, so the policy cover dimension is at most $d$. So, for Block MDP we have $\pcd \leq S$, and for Low Non-negative Rank MDP we have $\pcd \leq \dnnr$.

\section{Tabular MDP Sample Complexity Bound}

The actual bound available in \citet{azar2017minimax} is a regret bound. Given $N$ total episodes, as long as $N$ is sufficiently large (specifically, if $N \geq S^3 A$), and $H \leq S A$ (which is reasonable in non-trivial settings) they bound the total regret by
\begin{equation*}
    \textup{Regret}(N) \leq \widetilde\Ocal\Big( \sqrt{H^3 S A N} \log(\delta^{-1})^2 \Big) \,,
\end{equation*}
with probability at least $1-\delta$. Now, suppose we run UCB-VI over $N$ total episodes, for large $N$, and let $\hat \pi$ denote the average policy over all these episodes. Then, under the same high-probability event, the suboptimality of this average policy must be bounded by $\textup{Regret}(N)/N$. That is,
\begin{equation*}
    \SubOpt(\hat\pi; \Pi) \leq \widetilde\Ocal\Bigg( \log(\delta^{-1})^2 \sqrt{\frac{H^3 S A}{N}} \Bigg) \,.
\end{equation*}
Now, suppose we wish the sub-optimality to be bounded by some given $\epsilon \in (0,1)$. Then solving for $N$, this corresponds to
\begin{align*}
    &\widetilde\Ocal\Bigg( \log(\delta^{-1})^2 \sqrt{\frac{H^3 S A}{N}} \Bigg) \leq \epsilon \\
    &\iff N \geq \widetilde\Ocal\Big( H^3 S A \epsilon^{-2} \log(\delta^{-1})^4 \Big) \,.
\end{align*}
Then, noting that the above analysis is valid when $N \geq S^3 A$, we get a final bound of
\begin{equation*}
     N \geq \widetilde\Ocal\Big( H^3 S^3 A \epsilon^{-2} \log(\delta^{-1})^4 \Big) \,,
\end{equation*}
in order to ensure sub-optimality at most $\epsilon$, which is our presented sample complexity.

\section{Experimental Details}
\label{sec:exp}

We show the promise of $\algname$ on a rich observation MDP setting called Block MDP~\cite{du2019provably,misra2020kinematic}. In this setting, the agent receives observations generated from a latent state. Further, no two states can generate the same observation. This is an expressive MDP setting which can capture non-trivial practical problems. Our goal is to discuss the implementation details for $\algname$ and show that it can achieve optimal return, while not taking unsafe actions, and minimizing the calls to safety oracle. Note that all details of our environment and methodology will also be available in our code release, which can be used to fully reproduce all of our results, which is currently withheld to preserve anonymity

\paragraph{Environment.} For a given horizon $H$, an instance of the environment has $3H + 1$ states and $K=4$ actions. One of the action is a known safe action $\asafe$. These states are labeled as $\Scal = \{s_{1,0}\} \cup \{s_{i, h}\}_{i \in [4], h \in [H]}$.
 For a state $s_{i, h}$, the index $i$ denotes its type and $h$ denotes the level. We view states with index $i \in \{1, 2\}$ as normal states, while states index $i=3$ are considered safe states and those with $i=4$ are unsafe states. The agent never observes the state but instead receives an observation $x\sim q(\cdot \mid x)$ generated by an emission process $q$. The rich-observation setting implies that no two different states can generate the same observation. 

The agent starts deterministically in $s_{1, 0}$. All latent state transitions occur deterministically. After taking $h$ actions, the agent is in one of these four states: $\{s_{j, h}\}_{j \in [4]}$. Each environment instance has two action sequences $(a_1, \cdots, a_H)$ and $(a'_1, \cdots, a'_H)$ with $a_h\ne a'_h, a_h \ne \asafe$ and $a'_h\ne \asafe$, for all $h \in [H]$. We view the $a'_h$ as the \emph{unsafe action} and $a_h$ as the \emph{continue action} for time step $h$.

For any $i \in \{1, 2\}$ and $h \in [H]$, transitions in the normal state $s_{i, h}$ operate as follows: taking the continue action $a_h$ leads to the next normal state $s_{i, h+1}$ of same type, taking $\asafe$ leads to the safe state $s_{3, h+1}$, taking the unsafe action $a'_h$ leads to the unsafe state $s_{4, h+1}$, while the remaining action leads to the normal state normal state $s_{3 - i, h}$ of the other type. All actions in the safe state $s_{3, h}$ leads to the next safe state $s_{3, h+1}$. Finally, in the unsafe state $s_{4, h}$, taking any action except $\asafe$ leads to the next unsafe state $s_{4, h+1}$, while taking $\asafe$ leads to the safe state $s_{3, h+1}$. 

For a given transition $(s, a, s')$, the reward function $R(s, a, s')$ depends only on $s'$. For $s'=s_{i, h}$, the reward $r$ is defined as follows: if $i=1$ and $h < H$, then $r = \nicefrac{1}{H}$; for $i=1$ and $h=H+1$, $r = 2.0$, if $i=2$ then $r=-1$, if $i=3$ then $r=0$ and if $i=4$ then -1. The optimal return of 2.8 is achieved by staying on states with index 1.

An observation is stochastically generated from a state $s_{i, h}$ each time the agent visits the state. This is done by first concatenating two 1-sparse vectors encoding $i$ and $h$, and adding independent Gaussian noise to each dimension sampled from a 0 mean and 0.1 standard deviation. Let the resultant vector be $z'$ and it has $H + 5$ dimensions (4 dimensions to to encode $i$ and $H+1$ dimensions to encode $h$). We concatenate $z'$ with a 0 vector of size $2^{\log_2\lceil H + 5\rceil} - (H + 5)$. Let $z$ be the final vector and its dimensionality is given by $k=2^{\log_2\lceil H + 5\rceil}$. The observation is generated by multiplying $z$ with a Hadamard matrix of size $k\times k$ in order to mix its  dimensions. We use Sylvester's construction to create a Hadamard matrix $H_k$ of size $k\times k$. This is an inductive approach that works for $k = 2^l$ for some $l \in \NN$. It defines $H_1 = [1]$ and $H_{2n} = \left[\begin{array}{cc} H_n & H_n \\ H_n & -H_n\end{array}\right]$ for every $n \in \NN$. We borrowed the observation process from~\cite{misra2020kinematic}.

 In addition to the observation, the agent receives $(H+1)B$-dimensional safety features $\{\phi(s, a)\}_{a \in \Acal}$ for every action upon visiting a state. We assume that the safety function $f^\star$ is given by $f^\star(s, a) = w^T \phi(s, a) + b$ for some unknown $(w, b)$. In practice, these safety features can include readings from various safety devices attached to an agent (such as a LIDAR sensor, or temperature readings) which can be different from observation (e.g., camera image) which is used for dynamics. The agent receives a constant safe and unsafe feature for all states of type 3 and 4. States of type 2, have variance in safety features along all dimensions. Vising these states can help us learn safety function faster, but this comes at the cost of a negative reward associated with visiting these states. Finally, a state $s_{1, h}$ only have variance in safety features along the $\{(hB + 1, \cdots, hB + 1\}$ dimensions. Visiting $s_{1, h}$ gives higher reward but reveals the safety features more slowly. Therefore, an optimal agent must balance between optimizing reward and minimizing safety oracle calls by exploring safety features quickly.

\paragraph{$\algname$ Implementation.} We use Proximal Policy Optimization~\citep{schulman2017proximal} (PPO) as our blackbox RL algorithm. PPO is a popular empirical RL method and while not a PAC RL algorithm, is quite effective in practice for problems that do not require strategic exploration. We define the policy class $\Pi$ and the value network in PPO using a two-layer feed forward neural network with Leaky ReLU non-linearity.

We implement queries to region of disagreement by reducing it to linear programming. Formally, given a safety labeled dataset $\Dcal = \{(\phi(s_i, a_i), y_i)\}_{i=1}^n$, we solve for whether $\phi(s, a)$ is in $\rd^a(\Dcal)$, by first returning False if $a = \asafe$, and otherwise, solving the following two linear programs to compute the value of $c_{\max}$ and $c_{\min}$:
 \begin{align*}
     c_{\max}=&\max_{w, b} w^T \phi(s, a) + b, \quad \mbox{such that},\\
     &\forall i \in [d], \quad y_i(w^T \phi(s_i, a_i) + b_i) \ge 0,\\
     &\|w\|_{\infty} \le 1, |b| \le 1.
 \end{align*}
 and
  \begin{align*}
     c_{\min}=&\min_{w, b} w^T \phi(s, a) + b, \quad \mbox{such that},\\
     &\forall i \in [d], \quad y_i(w^T \phi(s_i, a_i) + b_i) \ge 0,\\
     &\|w\|_{\infty} \le 1, |b| \le 1.
 \end{align*}
 The given feature $\phi(s, a)$ is in $\rd(\Dcal)$ if and only if either $c_{\max} \ge 0$ and $c_{\min} < 0$, or $c_{\max} > 0$ and $c_{\min} \le 0$. 
 
We implement $\Pi(\Dcal)$ by mapping each policy $\pi \in \Pi$ to a \emph{known safe policy} $\pi_\Dcal$ and defining $\Pi(\Dcal) = \{\pi_\Dcal \mid \pi \in \Pi\}$. Let $(s, a) \in {\tt SurelySafe}(\Dcal)$ denote set membership in the set of surely safe state-action pairs. We define $\pi_\Dcal$ given $\pi$ as
 
 \begin{equation*}
     \pi_{\Dcal}(a \mid s) = \frac{\one\left\{(s, a) \in {\tt SurelySafe}(\Dcal)\right\}\pi(a \mid s)}{\sum_{a' \in \Acal}\one\left\{(s, a') \in {\tt SurelySafe}(\Dcal)\right\} \pi(a' \mid s)}.
 \end{equation*}
 
 When the agent visits a state $s$, we allow the agent to receive the safety features for all actions $\{\phi(s, a')\}_{a' \in \Acal}$. We test for set membership  $(s, a) \in {\tt SurelySafe}(\Dcal)$ by first returning True if $a = \asafe$, and otherwise, solving the linear programs described above using the feature $\phi(s, a)$ and returning True if and only if $c_{\max} \ge 0$ and $c_{\min} \ge 0$.

 Our implementation of $\algname$ closely follows~\pref{alg:ours}. One notable departure for efficiency is that we update the region of disagreement incrementally, rather than in batch.
 
 \paragraph{Hyperparameters.} We list hyperparameters that we use for $\algname$ for~\pref{tab:sabre_hyperparam} and for the PPO baseline in~\pref{tab:ppo_hyperparam}.
 
 \begin{table*}[!h]
     \centering
     \begin{tabular}{c|c}
        \hline 
          \textbf{Hyperparameter} & \textbf{Value}\\
        \hline 
          Learning Rate &  0.001\\
          Batch Size &  32\\
          N & 5 \\
          B & 1 \\
          m & 100 \\
          Num Eval & 500 \\
          Iterations of PPO & 1000 \\
          Gradient Clip & 20\\
          Number of PPO Updates & 10\\
          Ratio Clipping Coefficient & 0.1\\
          Entropy Coefficient & 0.01\\
          Iterations of PPO & 6500 \\
         \hline
     \end{tabular}
     \caption{Hyperparameters for $\algname$}
     \label{tab:sabre_hyperparam}
 \end{table*}
 
  \begin{table*}[!h]
     \centering
     \begin{tabular}{c|c}
        \hline 
          \textbf{Hyperparameter} & \textbf{Value}\\
        \hline 
          Learning Rate &  0.001\\
          Batch Size &  32\\
          Gradient Clip & 20\\
          Number of PPO Updates & 10\\
          Ratio Clipping Coefficient & 0.1\\
          Entropy Coefficient & 0.01\\
          Iterations of PPO & 6500 \\
         \hline
     \end{tabular}
     \caption{Hyperparameters for PPO}
     \label{tab:ppo_hyperparam}
 \end{table*}

\end{document}